\documentclass{article}

\usepackage[final]{neurips_2023}

\usepackage[utf8]{inputenc} % allow utf-8 input
\usepackage[T1]{fontenc}    % use 8-bit T1 fonts
\usepackage{hyperref}       % hyperlinks
\usepackage{url}            % simple URL typesetting
\usepackage{booktabs}       % professional-quality tables
\usepackage{amsfonts}       % blackboard math symbols
\usepackage{nicefrac}       % compact symbols for 1/2, etc.
\usepackage{microtype}      % microtypography
\usepackage{xcolor}         % colors
\usepackage[pdftex]{graphicx}
\usepackage{subcaption}     % subfigure
\usepackage{placeins}       % FloatBarrier
\usepackage{float}

\hypersetup{
    colorlinks,
    linkcolor={red!50!black},
    citecolor={gray},
    urlcolor={blue!80!black}
}

\title{Language Model Alignment with Elastic Reset}

\author{
  \hspace{-.7cm}Michael Noukhovitch\thanks{Correspondance to \texttt{michael.noukhovitch@umontreal.ca}} \\
  \hspace{-.7cm}Mila, Universit\'e de Montr\'eal\\
   \\
  \And
  \hspace{-.8cm}Samuel Lavoie \\
  \hspace{-.8cm}Mila, Universit\'e de Montr\'eal \\
  \AND
  Florian Strub \\
  Google Deepmind \\
  \And
  Aaron Courville \\
  Mila, Universit\'e de Montr\'eal \\
  Canada CIFAR AI Chair \\
}

\begin{document}

\maketitle

\begin{abstract}
Finetuning language models with reinforcement learning (RL), e.g. from human feedback (HF), is a prominent method for alignment. But optimizing against a reward model can improve on reward while degrading performance in other areas, a phenomenon known as reward hacking, alignment tax, or language drift. First, we argue that commonly-used test metrics are insufficient and instead measure how different algorithms tradeoff between reward and drift. The standard method modified the reward with a Kullback-Lieber (KL) penalty between the online and initial model. We propose \textbf{Elastic Reset}, a new algorithm that achieves higher reward with less drift without explicitly modifying the training objective. We periodically reset the online model to an exponentially moving average (EMA) of itself, then reset the EMA model to the initial model. Through the use of an EMA, our model recovers quickly after resets and achieves higher reward with less drift in the same number of steps. We demonstrate that fine-tuning language models with Elastic Reset leads to state-of-the-art performance on a small scale pivot-translation benchmark, outperforms all baselines in a medium-scale RLHF-like IMDB mock sentiment task and leads to a more performant and more aligned technical QA chatbot with LLaMA-7B. Code available at \href{https://github.com/mnoukhov/elastic-reset}{github.com/mnoukhov/elastic-reset}. %Inspired by recent work on iterated learning and resetting, we maintain an exponential moving average (EMA) of our model and every $n$ steps of training  1. reset our model to the EMA and 2. reset the EMA to the initial model. It is fast, simple, and does not impact training time. First demonstrate state-of-the-art performance on a small scale pivot-translation benchmark. Next we outperform all baselines in a medium-scale RLHF-like IMDB mock sentiment task. Finally, we apply Elastic Reset at large scale to align a Llama 7B model and create a more performant code assistant chatbot. 
\end{abstract}

\section{Introduction}

Dialogue agents that can effectively interpret and use language are a long-term challenge for NLP. The rise of large pretrained language models (LMs) \citep{brown_language_2020} made language model finetuning one of the most promising research directions to achieving capable dialogue agents \citep{bender_climbing_2020}. Recently, reinforcement learning (RL) has become a key ingredient of finetuning large LMs for interaction with humans \citep{ziegler_fine-tuning_2019,ouyang_training_2022,bai_training_2022}, notably shown in ChatGPT \citep{openai_chatgpt_2022}. A reward model is learned on the alignment objective, such as learned human preferences \citep[RLHF;][]{christiano_deep_2017,stiennon_learning_2020}, and the language model is finetuned to optimize the reward. But training on the RL objective moves the model away from its pretraining and can reduce performance on important benchmarks \citep{ouyang_training_2022} and even drifting away from natural language syntax and semantics \citep{lazaridou_multi-agent_2020}. 

``Language drift'' \citep{lee_countering_2019,lazaridou_multi-agent_2020}, ``alignment tax'' \citep{askell_general_2021}, ``reward model overoptimization'' \citep{gao_scaling_2022}, or LM-specific ``reward-hacking'' \citep{clark_faulty_2016} is inherent to RLHF. In the extreme case, models learn to achieve high reward by generating nonsense text that is unintelligible to humans \citep{lewis_deal_2017}. Methods to mitigate this issue range from re-running pretraining \citep{lowe_interaction_2021}, grounding in other modalities \citep{lee_countering_2019}, masking the LM generation \citep{ramamurthy_is_2022} and iterated learning \citep{lu_countering_2020}. But the standard, and by far most popular approach, adds a Kullback-Lieber (KL) divergence penalty to the reward in order to prevent the finetuned model from drifting too far from the pretrained model \citep{jaques_sequence_2017,jaques_way_2019,ziegler_fine-tuning_2019}. Still, all methods are insufficient over a large-enough training horizon so models are early-stopped before reaching a catastrophic level of drift. 

% fix up
\citet{gao_scaling_2022} find that achieving reward is proportional to drift from the initial model, but that not all drifts are equal. We wish to make small but effective changes that achieve high reward but maintain capabilities, yet auxiliary losses such as the KL penalty don't seem improve this tradeoff and only serve to slow down training \citep{gao_scaling_2022}. We posit that RLHF training requires a useful inductive bias that does not modify the training objective. Inspired by recent work in generalization for image classification \citep{zhou_fortuitous_2022}, sample-efficient RL \citep{nikishin_primacy_2022, doro_sample-efficient_2023}, and inducing compositional language \citep{li_ease--teaching_2019}, we propose to use resets. Similar to iterated learning \citep{kirby_spontaneous_2001}, iteratively resetting a model has been shown to reduce overfitting in language and RL scenarios \citep{rita_emergent_2022}. In this work, we show iteratively resetting a model also reduces drift while attaining equal or better reward than just a KL penalty. 

\begin{figure}
    \centering
    \includegraphics[width=0.9\linewidth]{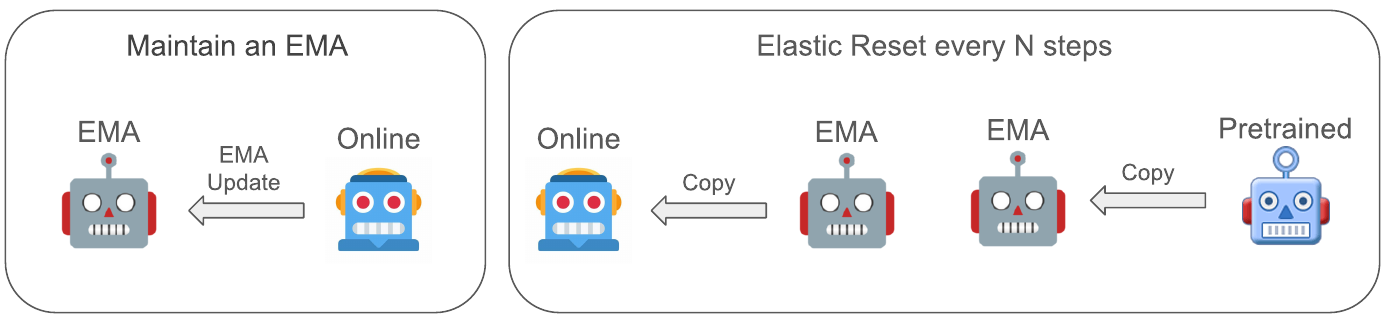}
    \caption{Elastic Reset. In actor-critic RL, we reset the policy but maintain the value function.}
    %For actor-critic methods, we only reset the policy but maintain the value function. }
    \label{fig:Elastic Reset}
\end{figure}

Unlike previous work in sample-efficient RL \citep{nikishin_primacy_2022}, RLHF is typically on-policy so it does not maintain a replay buffer with which to bootstrap learning after a reset. In lieu of a replay buffer, we reset the policy but maintain the value function.
Yet resetting the policy to its initial state can still cause a large drop in performance. So we propose resetting to a model in-between our online and initial state, specifically to an exponential moving average (EMA) of our online policy, as EMA has been shown to be highly performant \citep{caron_emerging_2021}. We still expect our EMA model to slowly drift, so we add a second step where we reset the EMA model to the initial model. We call this overall method \textbf{Elastic Reset} and illustrate it in Figure \ref{fig:Elastic Reset}. Elastic Reset is implemented on top of regular RL methods such as REINFORCE \citep{williams_simple_1992} or PPO \citep{schulman_proximal_2017}.

%Since EMA updates are very fast and negiligable compared to regular training time, the method is efficient and  practically no extra training time.

First, we test our method on a small scale task: pivot translation with a transformer. In this classic benchmark for drift, we outperform all previous baselines and demonstrate state-of-the-art performance. Next, we re-evaluate how performance is measured in the field and argue for a metric of how each method trades off performance vs drift. We propose the Pareto Frontier Graph, a graphical measure that illuminates the trade-off between performance and drift and demonstrate that Elastic Reset dominates the baselines against this trade-off. Then, we scale up slightly to GPT2 and work on a popular task closer to RLHF, IMDB mock sentiment. Comparing to all baseline methods, we again show state-of-the-art performance on the benchmark. Through ablations, we show that Elastic Reset is robust to choices of hyperparameters, even more so than baselines. Finally, we scale up even more to true RLHF finetuning of Llama-7B in order to create a helpful technical QA chatbot using a StackExchange dataset. We again outperform the baseline, demonstrating how Elastic Reset mitigates the alignment tax while better optimizing the human feedback reward.  

\section{Related Work}

``It is often difficult or infeasible to capture exactly what we want an agent to do, and as a result we frequently end up using imperfect but easily measured proxies'' \citep{clark_faulty_2016}. In RL, this proxy is how we construct our reward and the consequence can be ``reward-hacking'' \citep{clark_faulty_2016}; an agent optimizes the reward but does not accomplishing the meaningful task. RLHF aims to align an agent with human preferences while maintaining the capabilities of the pretrained model, but uses a learned reward model as a proxy of human preferences \citep{christiano_deep_2017,ziegler_fine-tuning_2019}. This can lead to LMs that optimize a reward model but degrade in performance on general NLP benchmarks \citep{askell_general_2021}, overfit the reward model and do not generalize to true human preferences \citep{bai_training_2022,gao_scaling_2022}, or latch onto confounding factors hidden in the reward \citep{stiennon_learning_2020}. These effects are exacerbated if the reward model is updated during training such as in iterated RLHF \citep{bai_training_2022} or related setups such as emergent communication \citep{lazaridou_emergent_2020}, end-to-end dialogue \citep{lewis_deal_2017}, learning RL policies through latent language \citep{andreas_learning_2018}, and pivot translation \citep{utiyama_comparison_2007}. There, the phenomenon is known as ``language drift'' \citep{lee_countering_2019} and can lead to incoherent and unnatural linguistic outputs \citet{lewis_deal_2017}.

This phenomenon is inherent to RLHF. \citet{gao_scaling_2022} show that improvement on alignment / reward is proportional to drift from the initial model, but also find that different methods and design choices achieve different proportions of performance to drift. Therefore, a major challenge of RLHF is how to learn the reward in such a way as to minimize the drift, alignment tax, and reward-hacking. The standard approach used in most RLHF is to incorporate a KL penalty between the training language model and some fixed model \citep{jaques_way_2019, ziegler_fine-tuning_2019, stiennon_learning_2020, ouyang_training_2022, steinert-threlkeld_emergent_2022, bai_training_2022}, usually the initial, pretrained model. Less common is to add the original pretraining task to the finetuning objective (termed S2P by \citet{lowe_interaction_2021}) but this can be compute-intensive and requires maintaining the pretraining data which may be even more expensive for larger models \citep{brown_language_2020}.
% In general, we require a method that does not rely on pretraining data and relative compute efficiency if it is to be implemented at scale. 
On small-scale pivot-translation, \citet{lu_countering_2020} propose iterated learning with student-teacher distillation but it too is relatively compute intensive. Recently, \citet{ramamurthy_is_2022} propose to maintain a delayed masking model and mask the LM to output only the top-$p$ tokens. Elastic Reset takes inspiration from both of these, using an iterated process and maintaining an EMA model. Apart from better performance, our method is more space efficient and maintains the EMA on CPU whereas both other methods require maintaining an extra model on GPU. It is also more compute efficient as resetting weights and EMA updates are very cheap operations, whereas \citet{lu_countering_2020} requires a long distillation phase and \citet{ramamurthy_is_2022} requires an extra forward pass with the masking model. There exist less-popular methods that have been applied to similar issues in RL: prompt-tuning \citep{singh_know_2022}, using a fixed model to generate many options and re-ranking using a reward model \citep{lazaridou_multi-agent_2020, meta_fair_diplomacy_team_human-level_2022}, and grounding the output in a separate modality \citep{lee_countering_2019} but none have been used for RLHF or at scale.

Our method is inspired by recent works that leverage resets for single agent RL \citep{nikishin_primacy_2022, doro_sample-efficient_2023}, image classification \citep{zhou_fortuitous_2022}, and emergent communication \citep{rita_emergent_2022}. Those works generally train from scratch and reset to random initializations in order to improve generalization. Our scenario requires resetting to pretrained models and focuses on improving the tradeoff between performance and drift from this pretrained model. Elastic Reset can be seen as an on-policy alternative to \citeauthor{nikishin_primacy_2022}'s \citeyearpar{nikishin_primacy_2022} 
off-policy resets; whereas they maintain the old replay buffer, we maintain the value model and an EMA of our policy.

Finally, the pretrain-then-RL-finetune setup with the goal of maintaining pretrained knowledge can be seen as a two-step, RL-specific instance of continual learning and therefore language drift has links to catastrophic forgetting \citep{mccloskey_catastrophic_1989}. There is a clear similarity between mitigation methods: rehearsal \citep{robins_catastrophic_1995} or experience replay \citep{rolnick_experience_2019} is equivalent to multitasking with the pretraining objective \citep{lowe_interaction_2021} and weight-update regularization \citep{kirkpatrick_overcoming_2017} has similarities to KL regularization \citep{jaques_way_2019}.

% This can equivalently be seen as variation inference: approximating a Bayesian posterior which specifies how to update a prior LM to conform with evidence provided by the reward function.

% Language drift is also present when learning policies through latent language in RL \citep{andreas_learning_2018} and similarly, multi-tasking has been shown to be effective there \citep{jacob_multitasking_2021}. 
% The pretrain-then-finetune setup with the goal of maintaining pretrained knowledge can be seen as a specific instance of continual learning and therefore language drift has links to catastrophic forgetting \cite{mccloskey_catastrophic_1989}. There is a clear similarity between mitigation methods: rehearsal \citep{robins_catastrophic_1995} or experience replay \citep{rolnick_experience_2019} is equivalent to multitasking with the pretraining objective \citep{lowe_interaction_2021} and weight-update regularization \citep{kirkpatrick_overcoming_2017} has similarities to KL regularization. 

\section{Elastic Reset}
% Give a formal definition of RLHF, LMs + RL?

The standard method against drift is a KL penalty, generally between the learning policy $\theta$ and the initial, pretrained model $\theta_0$. It is calculated empirically over the minibatch of training inputs $x$ and outputs $y$ and used as an auxiliary reward with coefficient $\beta$ on top of the regular reward model $r$

\begin{equation}
    R(x,y) = r(x,y) - \beta \log \frac{\pi_\theta(y|x)}{\pi_{\theta_0} (y|x)}
\end{equation}

For Elastic Reset, we maintain an exponential moving average $\bar \theta$ of our learning model $\theta$ and choose a decay hyperparameter parameter $\eta$. We initialize $\bar \theta \gets \theta_0$ and after every online model step, we update our EMA model $\bar \theta \gets (1 - \eta) \theta + \eta \bar \theta$. Every $n$ steps, Elastic Reset sets the online model to the EMA model $\theta \gets \bar \theta$ and sets the EMA model to the initial model $\bar \theta \gets \theta_0$. As with other methods, Elastic Reset can be easily combined with a KL penalty.
%, we refer to this as Elastic Reset + KL.

% As we later show, setting to an EMA model on its own is effective, but is further boosted by changing the KL from a pretrained, to an EMA model. 
% The KL reduces the LM's drift from the EMA model which guarantees that the EMA is closer in parameter space. This, in turn, reduces the difference in performance before and after resetting to the EMA model.

\begin{figure}[t]
    \centering
     \begin{subfigure}[b]{0.49\linewidth}
         \centering
         \includegraphics[width=\linewidth]{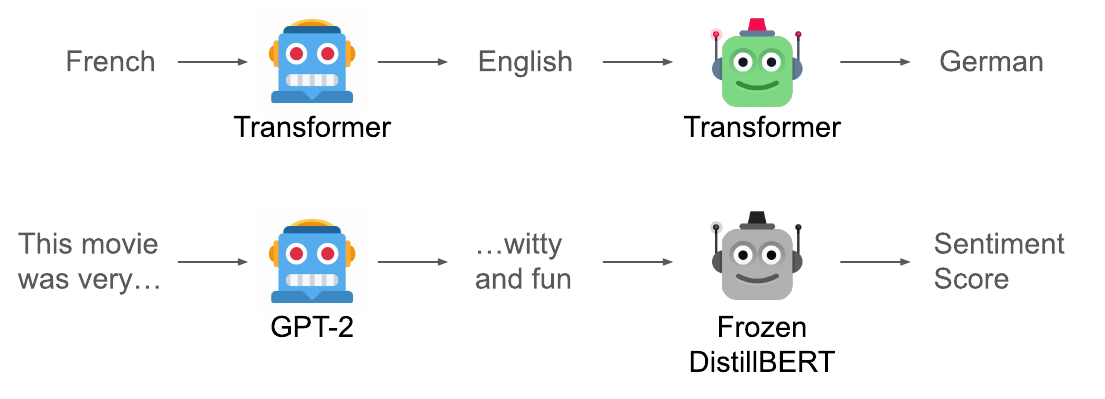}
         % \caption{Language Drift}
         % \label{fig:f}
     \end{subfigure}
     \hfill
     \begin{subfigure}[b]{0.49\linewidth}
         \centering
         \raisebox{0.5cm}{\includegraphics[width=\linewidth]{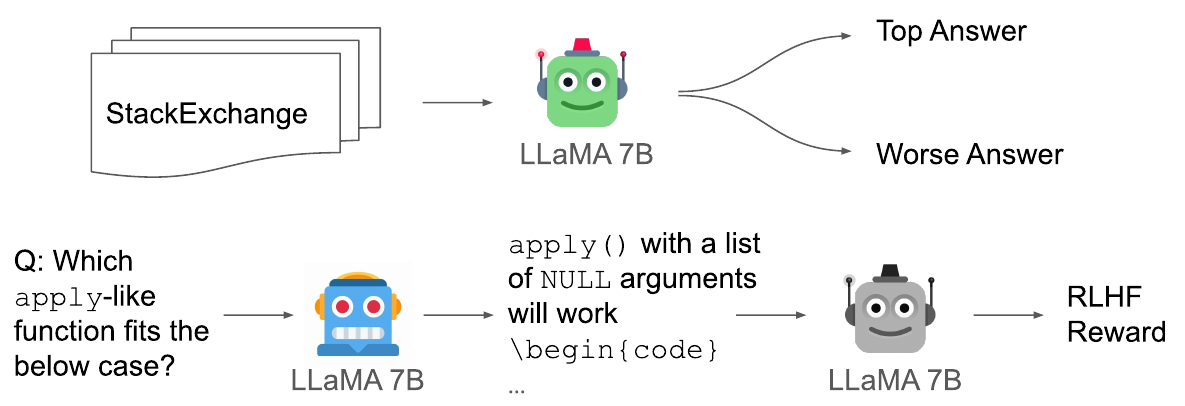}}
         % \caption{Pareto Frontier Graph}
         % \label{fig:translation-pareto}
     \end{subfigure}
    \caption{The Translation Game (top left), IMDB mock sentiment task (bottom left), and StackLLaMA (right). We show all RL finetuning setups and StackLLaMA's reward modelling (top right). 
    % The reward model in the IMDB task is pretrained then frozen during finetuning, whereas the reward model of the Translation Game is dynamic and updated during finetuning. 
    }
    \label{fig:tasks}
\end{figure}

\section{Translation Game: Careful Comparison to SOTA}

\paragraph{Setup} We first investigate the pivot-translation benchmark of \citet{lee_countering_2019}, which was previously popular for small-scale methods countering drift. Two translation models, French to English (FR$\to$EN) and English to German (EN$\to$DE), are pretrained on IWSLT \citep{cettolo_wit3_2012}. Then, the models are finetuned on translating French to German through English (FR$\to$EN$\to$DE) but given only paired French and German data from Multi30k \citep{elliott_multi30k_2016, elliott_findings_2017} as shown in Figure \ref{fig:tasks}. The models are not given English at finetune-time so the challenge is optimizing FR$\to$DE while maintaining fluency in the intermediate English. Whereas larger benchmarks have only proxies for drift, we can exactly measure the performance degradation in our setup with the standard translation metric BLEU on a held-out FR$\to$EN validation set. Similarly, we measure success on the task with the FR$\to$EN$\to$DE BLEU score. Each model is an encoder-decoder Transformer \citep{vaswani_attention_2017} with 6 layers and all experimental details are available in Appendix \ref{app:translation-game}. 
% From the perspective of a sender-receiver game \citep{lewis_convention:_1969} we consider FR$\to$EN to be our sender and EN$\to$DE to be our receiver whereas in the perspective of RL+LM, FR$\to$EN is our LM and EN$\to$DE is an online learning reward model.   

\paragraph{Baselines} The EN$\to$DE reward model is simply trained using cross-entropy between predicted and true DE. Our lower-bound baseline is \textsc{Frozen English}, we freeze the FR$\to$EN model and only update the EN$\to$DE model. This models is guaranteed not to drift, but also cannot reach the best possible performance. For that, we need to update FR$\to$EN by backpropogating through the discrete EN tokens. We follow \citet{lee_countering_2019} and train FR$\to$EN using REINFORCE \citep{williams_simple_1992} to estimate the gradient. As our base model, we combine \textsc{REINFORCE} with an exponentially moving baseline and, as with previous work, add a loss for entropy regularization.

When both FR$\to$EN and EN$\to$DE are being updated, we tend to see reasonably large drift and we compare to the best previous methods that counter it on this benchmark. We follow \citet{lee_countering_2019} to simulate the standard KL penalty method, \textsc{KL penalty} by training an LSTM LM on IWSLT English text and adding a KL penalty with $\beta = 0.05$ to regularize the FR$\to$EN model. \textsc{Multitask} learning, re-training,  or S2P \citep{lowe_interaction_2021}, adds the supervised FR$\to$EN objective on IWSLT pretraining data as an auxiliary task for the FR$\to$EN model. Finally, we implement Seeded Iterated Learning \citep[\textsc{SIL};][]{lu_countering_2020}, which alternates between $n$ finetuning steps and $m$ steps of teacher-student distillation.  FR$\to$EN and EN$\to$DE ``teacher'' models are finetuned on the translation game, then each distills knowledge into ``student'' model of itself, and finally the students are initialized as teachers for the next iteration.  \textsc{Elastic Reset} is implemented on top of REINFORCE with a very minimal KL penalty $\beta = 0.001$ and uses an EMA decay $\eta = 0.99$. We run all models for 50k updates and reset every 23k steps to get 2 resets / 3 iterations within a run. Hyperparameters may differ between methods, e.g. $\beta$, because we used a minimal search to find the best hyperparameters for each method.

\paragraph{Experiments}

\begin{figure}
     \centering
      \begin{subfigure}[b]{0.32\linewidth}
         \centering
         \includegraphics[width=\linewidth]{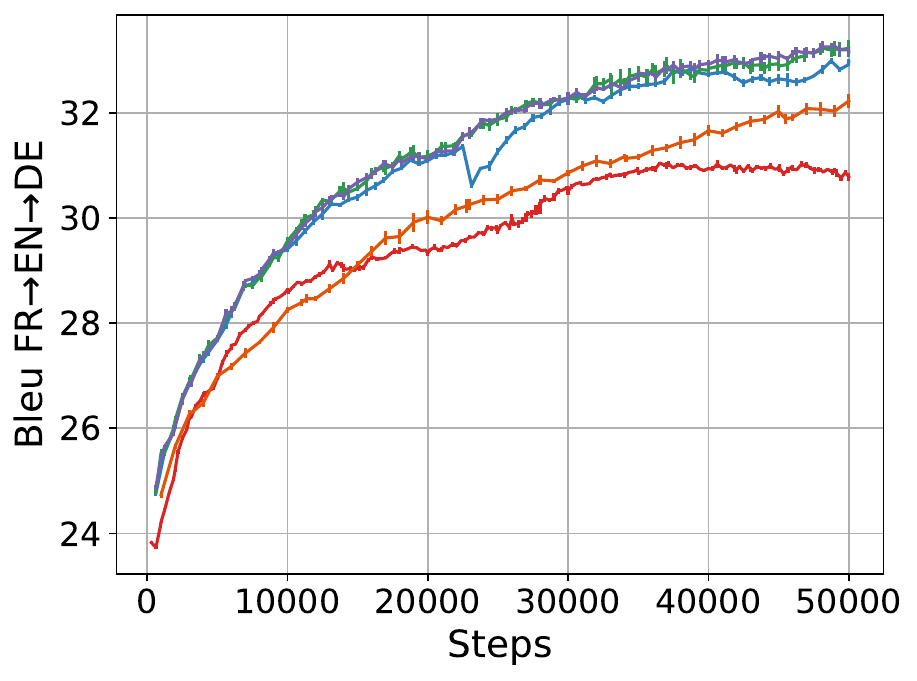}
         \caption{Task Performance}
         \label{fig:fr-en-de}
     \end{subfigure}
     \begin{subfigure}[b]{0.32\linewidth}
         \centering
         \includegraphics[width=\linewidth]{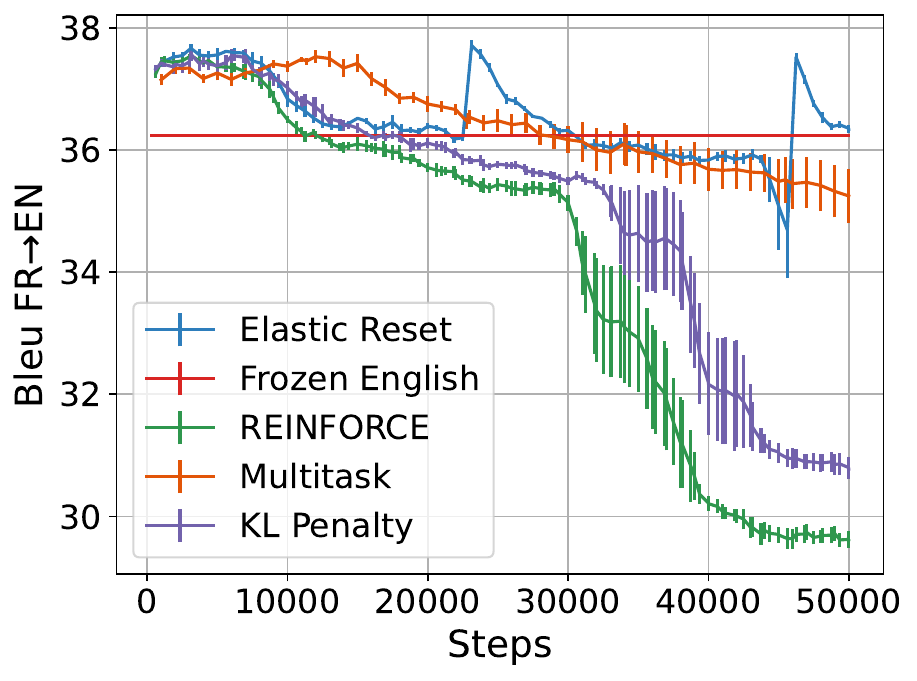}
         \caption{Language Drift}
         \label{fig:fr-en}
     \end{subfigure}
     \begin{subfigure}[b]{0.32\linewidth}
         \centering
         \includegraphics[width=\linewidth]{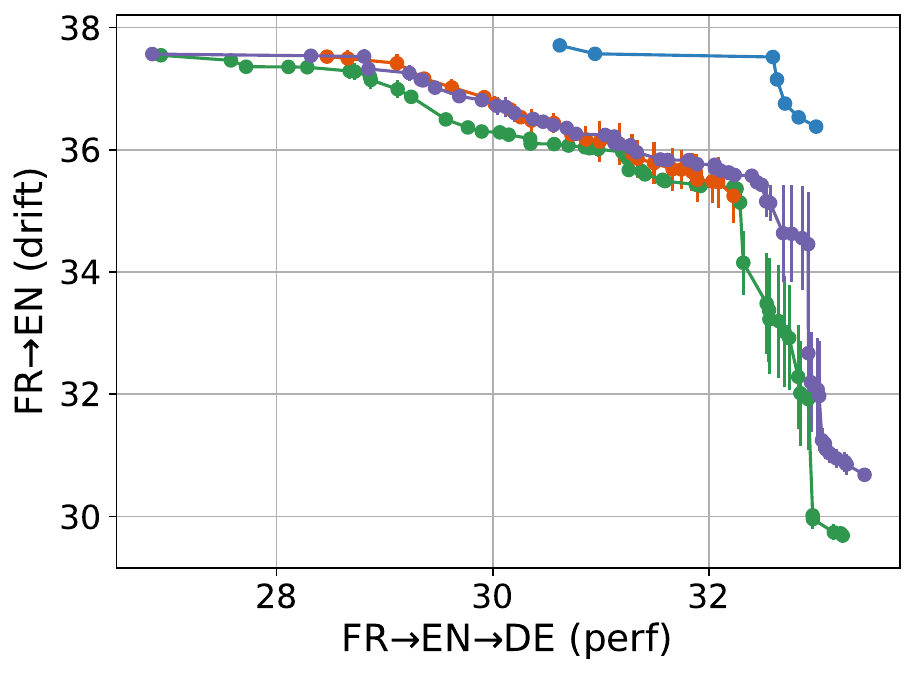}
         \caption{Pareto Frontier Graph}
         \label{fig:translation-pareto}
     \end{subfigure}
    \caption{Comparing Elastic Reset to all baseline methods on the Translation Game. We measure (a) Task Performance with FR$\to$EN$\to$DE BLEU and (b) Language Drift with FR$\to$EN BLEU, on the validation set during finetuning. We plot the mean and standard error over 5 seeds. To compare how methods trade off the two metrics, we plot (c) the best achieved drift vs task performance.}
    \label{fig:translation-game}
    \vskip -0.1in
\end{figure}

For each method, we run 5 seeds and plot the validation scores over training for the end-to-end task score, FR$\to$EN$\to$DE BLEU, and drift score, FR$\to$EN BLEU, in Figures \ref{fig:fr-en}, \ref{fig:fr-en-de} respectively. Following \citet{lee_countering_2019}, we also show the final validation score in Table \ref{tab:translation-game}. As a sanity check, \textsc{Frozen English} does not drift but also does not achieve a very high task performance. In line with previous results \citep{lu_countering_2020}, all learning models initially improve FR$\to$EN performance, likely because models are quickly, semi-supervised adapting from their pretraining (IWSLT) to the distribution of the finetune dataset (Multi30k). Afterwards, they start to overfit on their objective and FR$\to$EN performance degrades. \textsc{REINFORCE} achieves the best possible task performance but drifts significantly. Despite extensive hyperparameter tuning and correspondence with the original authors, \textsc{SIL} does not manage to outperform the REINFORCE baseline so we exclude it from the figures for visual clarity but show values in Table \ref{tab:translation-game} as well as full results in Appendix \ref{app:translation-game}. In line with previous work \citep{lee_countering_2019,lu_countering_2020}, we find that \textsc{multitask} and \textsc{KL penalty} are both beneficial to reducing drift, but both represent a tradeoff. Whereas \textsc{multitask} strongly reduces drift, it does not achieve a high task score. In contrast, \textsc{KL penalty} achieves a high task score but drifts quite drastically. Elastic Reset achieves nearly the best possible task score while maintaining the same drift score as the initial model. Visually, we see that our method track the baselines until the reset at 23k steps. After the reset, we see a slight performance drop but also a big jump back in terms of FR$\to$EN drift. While the task performance recovers within 5k steps, the drift performance does not degrade to previous levels. For the second reset, the EMA model is slightly more drifted and so the reset is less pronounced for both task and drift, leading to faster task recovery but slightly more drift. Overall, Elastic Reset shows state-of-the-art results on the benchmark and outperforms all previous small-scale methods. 

\begin{table}[t]
\caption{Translation Game final validation scores }
\label{tab:translation-game}
% \vskip -0.2in
\begin{center}
\begin{small}
\begin{sc}
\begin{tabular}{lcc}
\toprule
 & $\uparrow$ FR$\to$EN$\to$DE & $\uparrow$ FR$\to$EN \\
\midrule
Frozen English   & 30.8$\pm$0.2 & \textbf{36.3$\pm$0.1}  \\
REINFORCE       & \textbf{33.2$\pm$0.3} & 29.6$\pm$0.3 \\
+ SIL           & 28.2$\pm$0.4 & 27.3$\pm$4.4 \\
+ Multitask (S2P) & 32.2$\pm$0.3  & 35.2$\pm$1.0  \\
+ KL penalty & \textbf{33.2$\pm$0.2}  & 30.8$\pm$0.4  \\
+ \textbf{Elastic Reset}       & 32.9$\pm$0.1 & \textbf{36.3$\pm$0.1} \\
% + EMAreset     & 33.0$\pm$0.1 & \textbf{36.3$\pm$0.1} \\
\bottomrule
\end{tabular}
\end{sc}
\end{small}
\end{center}
\vskip -0.1in
\end{table}

\section{Pareto Frontier Graph}

Simply evaluating validation curves side-by-side or looking at a table of final scores, it can be unclear which method is better if one drifts less but the other achieves a higher reward e.g. \textsc{Multitask} vs \textsc{KL penalty} in Table \ref{tab:translation-game}. Previous work on this \citep{lee_countering_2019} and other benchmarks \citep{ramamurthy_is_2022} compare methods using simple point-estimates after training for a specific number of epochs. But this number of epochs is quite arbitrary as models never fully converge to a reward, they are early-stopped such that drift is not catastrophic. Since different setups may admit different levels of drift, we believe that evaluation should reflect the continuous tradeoff between task and drift. We extend \citet{ziegler_fine-tuning_2019}, and create a pareto frontier graph to plot each method's achieved task score vs drift metric on the validation set over training. We believe practioners will wish to choose the best model for some given task performance so, contrary to \citet{ziegler_fine-tuning_2019}, we plot the task score on x-axis and drift score on the y-axis. Improvement on a drift metric can either mean lower scores (perplexity) or higher scores (BLEU) so we always plot task score as increasing from bottom to top such that, graphically, a better method will functionally dominate a worse method. We plot the best achieved reward vs drift over all validation steps for the Translation Game in Figure \ref{fig:translation-pareto}. Not only does Elastic Reset outperform the baselines at the final validation score, but it functionally dominates such that it is the best method for all levels of task performance it achieves.

\section{IMDB Mock Sentiment: Ablation Study for RLHF}
\label{sec:imdb}

\paragraph{Setup} Next, we scale to a larger benchmark that more closely approximates the standard RLHF setup. We use the recently released GRUE benchmark for RL training of LMs \citep{ramamurthy_is_2022} and use IMDB mock sentiment \citep{ziegler_fine-tuning_2019}, the main task where language models are susceptible to reward-hacking, shown in Figure \ref{fig:tasks}. The goal is to complete an IMDB movie review with as positive a sentiment as possible. The baseline LM is GPT-2 \citep{radford_language_2019} with 117M parameters further pretrained on the IMDB domain \citep{maas_learning_2011}. We learn a DistilBERT \citep{sanh_distilbert_2020} reward model on IMDB to output a sentiment score between 0 (negative) and 1 (positive). We then train our GPT-2 LM to complete different IMDB reviews while maximizing the sentiment reward. Following \citet{ramamurthy_is_2022}, we measure reward-hacking / drift with our model's perplexity on the true IMDB data. If we consider knowledge of the IMDB data as a useful capability, then our initial model was finetuned on IMDB to maximize log-probability, i.e. minimize perplexity, and has the maximum capabilites. We measure divergence from the initial model, and decrease in capabilities, by the increase in our trained model's perplexity on ground truth IMDB data. In contrast to the previous task, a lower perplexity score corresponds to less drift.

\paragraph{Baselines} Our main baseline is \textsc{PPO} \citep{schulman_proximal_2017} with \citet{ziegler_fine-tuning_2019} modifications for RLHF training, specifically adding a KL penalty with the frozen initial model (equivalent to \textsc{KL with pretrained}) and dynamically decaying the coefficient $\beta$ over training. To further increase stability, Generalized Advantage Estimation \citep{schulman_trust_2015} is used for the advantage estimator. We also compare to \textsc{NLPO} \citep{ramamurthy_is_2022}, a recent method that extends PPO with a masking model to counteract drift. The masking model is initialized to the pretrained model and recieves delayed updates; it is set to the online model every $n$ steps. During training, the online model's output probabilities are restricted to the mask model's top $p$ tokens. We use the RL4LMs library \cite{ramamurthy_is_2022} and their default hyperparameters for both PPO and NLPO e.g. $\beta = 0.1$. We implement \textsc{Elastic Reset} on top of PPO with an EMA decay rate of $0.995$ and greatly reduce the KL coefficient $\beta = 0.001$ to allow the model to drift more, then reset every 17 epochs such that we get two resets / three iterations during our training. 

\begin{figure}[t]
     \centering
      \begin{subfigure}[b]{.32\linewidth}
         \centering
         \includegraphics[width=\linewidth]{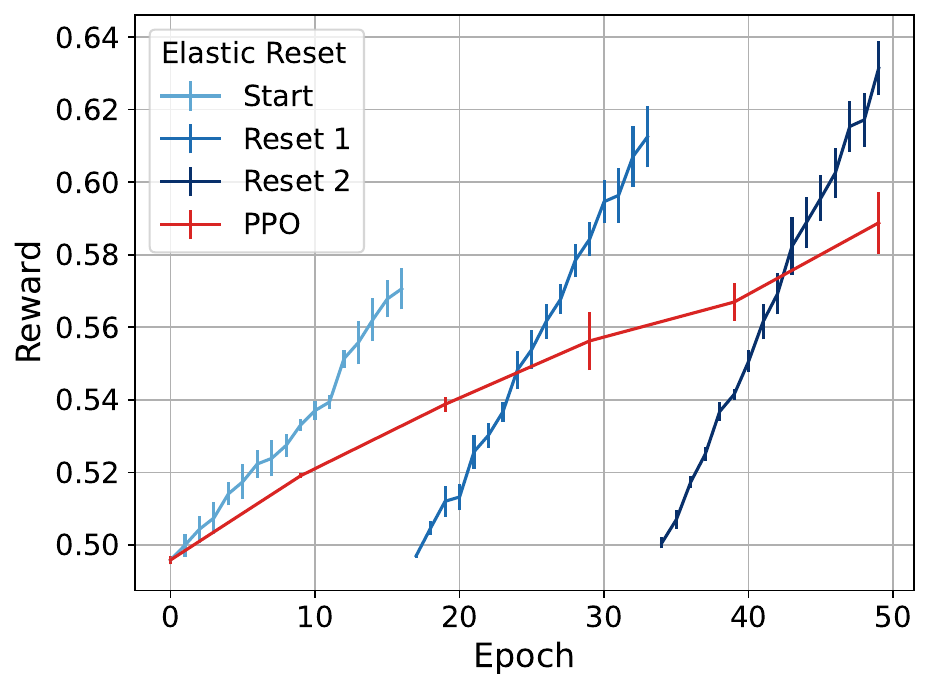}
         \caption{Task Performance}
     \end{subfigure}
     \begin{subfigure}[b]{.32\linewidth}
         \centering
         \includegraphics[width=\linewidth]{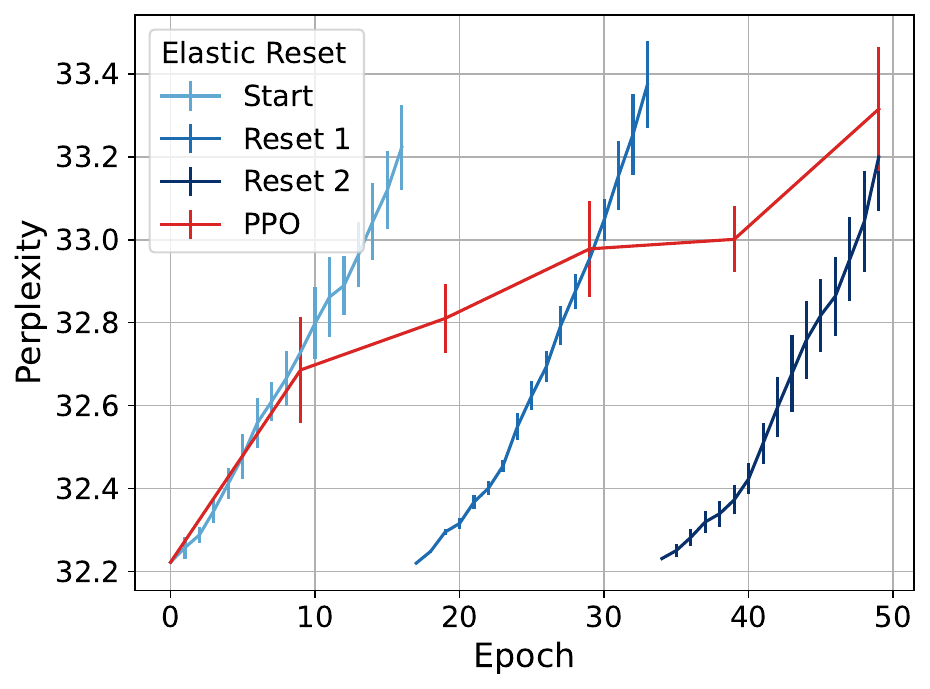}
         \caption{Language Drift}
     \end{subfigure}
     \begin{subfigure}[b]{.32\linewidth}
         \centering
         \includegraphics[width=\linewidth]{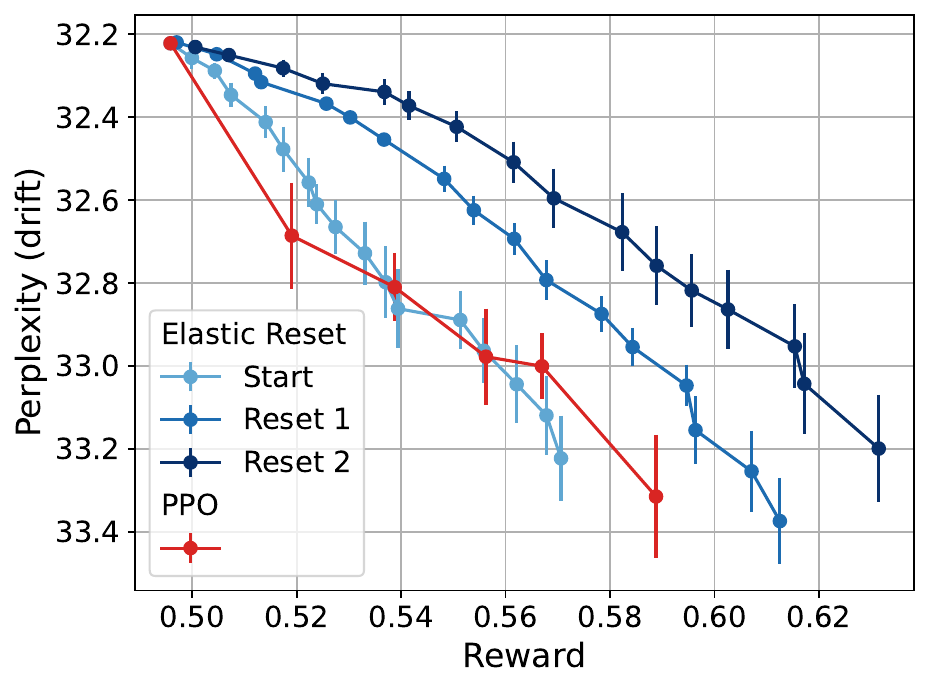}
         \caption{Pareto Frontier Graph}
     \end{subfigure}
    \caption{Plotting PPO vs Elastic Reset on IMDB but splitting the results visually between resets.  We measure (a) Language Drift and (b) Task Performance via Semantic Score on the validation set over finetuning. All methods also include a KL penalty. We plot mean and standard error across 5 seeds.}
    \label{fig:imdb}
\end{figure}

\begin{table}[t]
\caption{IMDB mock sentiment final test scores}
\label{tab:imdb}
\begin{center}
\begin{sc}
% \resizebox{0.5\linewidth}{!}{
\begin{tabular}{lcc}
\toprule
 & $\uparrow$ Sentiment & $\downarrow$ Perplexity \\
\midrule
Zero-shot & .489$\pm$0.01 & 32.45$\pm$0.13  \\
PPO  & .596$\pm$0.02  & 33.45$\pm$0.40  \\
NLPO & .558$\pm$0.06 & 33.12$\pm$0.74 \\
\textbf{Elastic Reset}  &  .611$\pm$0.02    & 33.32$\pm$0.23 \\
% PPO + KL EMA     & .620$\pm$0.05 & 33.73$\pm$0.45 \\
% + \textbf{Reset to EMA}      & \textbf{.636$\pm$0.03} & \textbf{33.65$\pm$0.31} \\
% + \textbf{Elastic Reset}     & \textbf{.611$\pm$0.02} & \textbf{33.32$\pm$0.23} \\ 
\bottomrule 
\end{tabular}
% }
\end{sc}
% \scriptsize	
%\end{small}
\end{center}
\vskip -0.1in
\end{table}

\paragraph{Experiments} We run all experiments for 5 seeds and report mean and standard error on our validation set for our reward, DistilBERT sentiment, and our drift score, perplexity. Following \citet{ramamurthy_is_2022}, we run for 50 epochs (equivalent to 64k updates) and show our results in Figure \ref{fig:imdb}. To make our resets more visible, we plot validation scores every epoch for Elastic Reset. Since the benchmark provides a test set as well, we compare all final models in Table \ref{tab:imdb}. The PPO baseline performs quite well because it already includes a KL with the pretrained model. We find NLPO performs similarly to PPO, so we relegate NLPO results to Appendix~\ref{app:imdb}. Results from the original NLPO paper were stronger \citep{ramamurthy_is_2022} but our reproduced numbers and curves were confirmed by the original authors \citep{ammanabrolu_re_2023}. Elastic Reset achieves better semantic scores much faster by using a smaller KL penalty coefficient ($0.001$ vs  PPO $0.1$) but also drifts more to achieve them. As with the previous task, this drift is then mitigated by the reset and we see semantic task score improve in relation to drift over the iterations. Looking at the pareto graph in Figure~\ref{fig:imdb-pareto}, we see that Elastic Reset far outpaces the baselines and provides a better tradeoff of reward vs drift for every reward.  
% We also show positive results with the other baseline model, a supervised fintuned GPT-2 in Appendix \ref{app:imdb-sup}.

\begin{figure}
     \centering
     \begin{subfigure}[b]{0.4\linewidth}
         \centering
         \includegraphics[width=\linewidth]{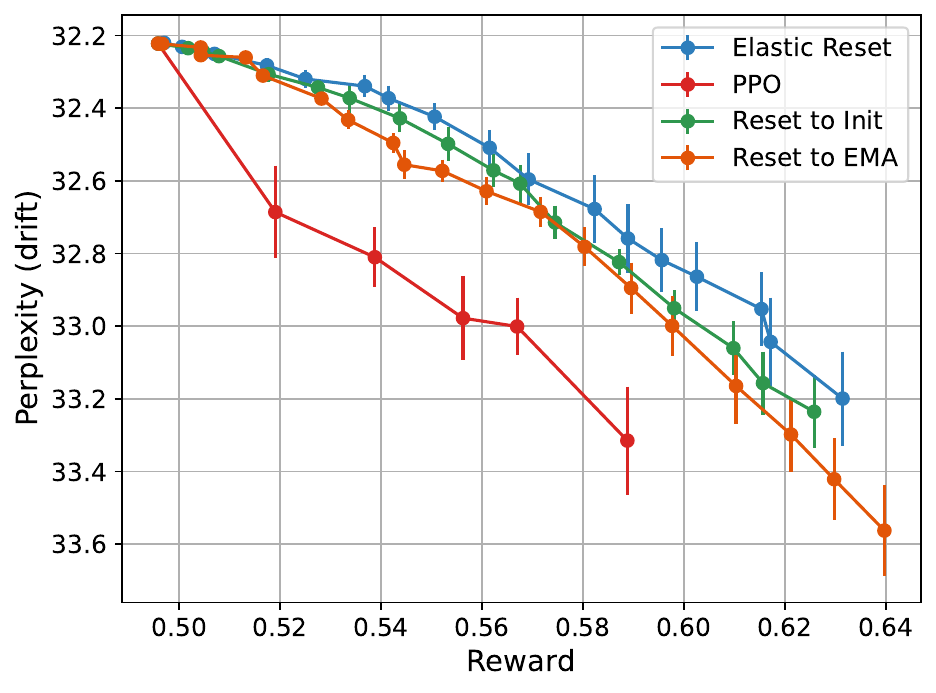}
         \caption{Reset Ablation}
         \label{fig:imdb-ablation-reset}
     \end{subfigure}
     \hspace{1cm}
      \begin{subfigure}[b]{0.4\linewidth}
         \centering
         \includegraphics[width=\linewidth]{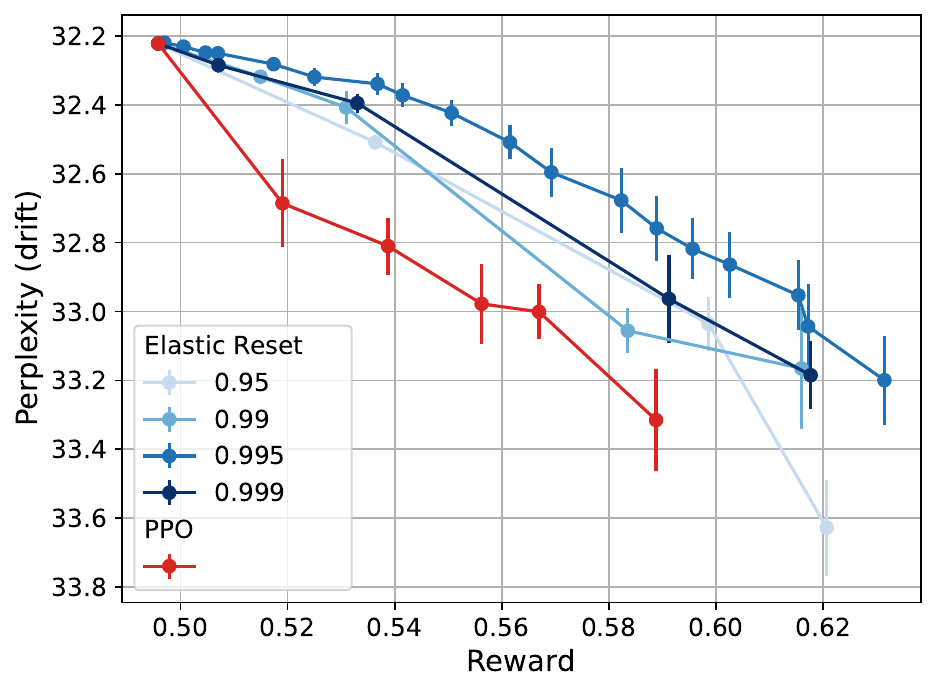}
         \caption{EMA Decay $\eta$ Ablation}
         \label{fig:imdb-ablation-decay}
     \end{subfigure}
      \begin{subfigure}[b]{0.4\linewidth}
         \centering
         \includegraphics[width=\linewidth]{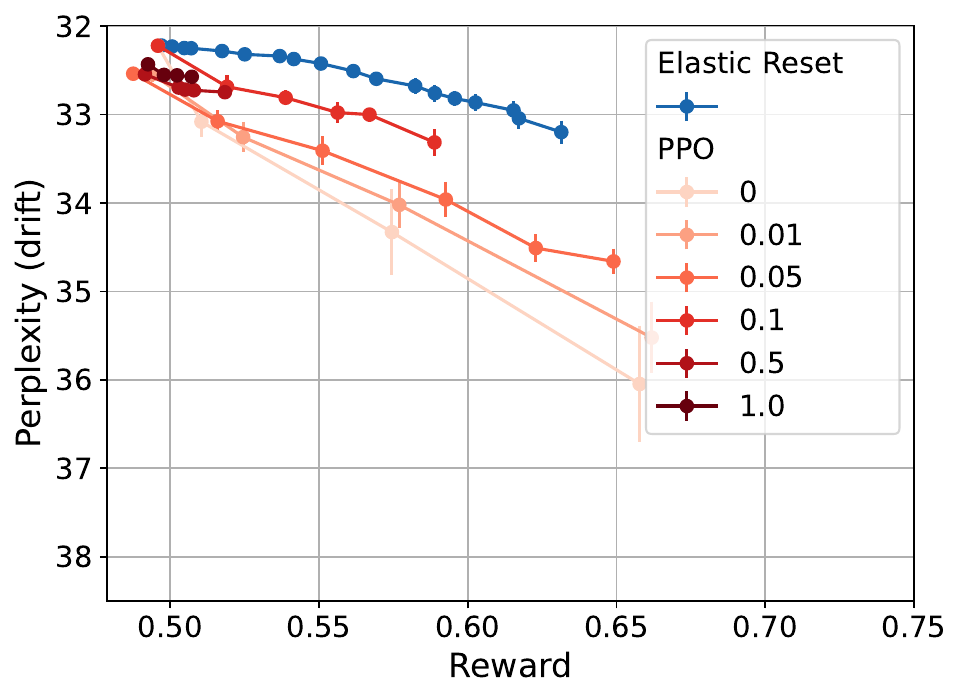}
         \caption{KL $\beta$ Ablation for PPO}
         \label{fig:imdb-klablation-ppo}
     \end{subfigure}
     \hspace{1cm}
      \begin{subfigure}[b]{0.4\linewidth}
         \centering
         \includegraphics[width=\linewidth]{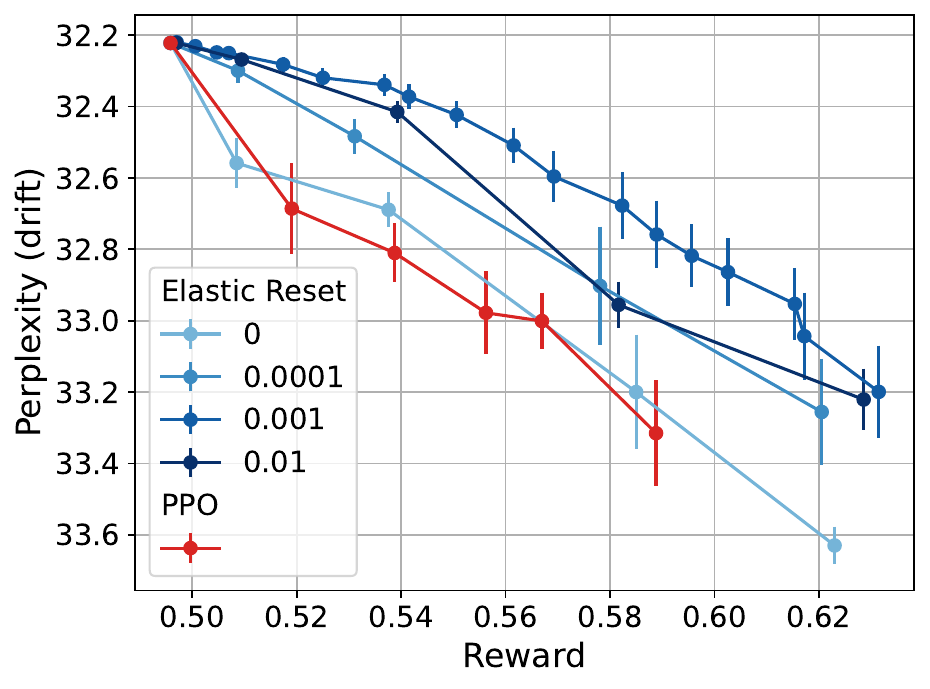}
         \caption{KL $\beta$ Ablation for Elastic Reset}
         \label{fig:imdb-klablation-emaset}
     \end{subfigure}
    \caption{Ablating Elastic Reset on the IMDB mock sentiment task. We plot pareto graphs using mean and standard error across 5 seeds.}
    \label{fig:imdb-ablations}
\end{figure}

\paragraph{Ablations}
We empirically investigate our method through ablations. Throughout this section we run experiments on IMDB with the same hyperparameters, unless otherwise mentioned. For brevity, we plot only the pareto graphs but include all other graphs in Appendix~\ref{app:ablations} along with these same ablation experiments for the Translation Game, with similar results.

To investigate the source of improvement in Elastic Reset, we ablate the two resets: online to EMA, and EMA to initial model. We discard the second reset to get \texttt{Reset to EMA}: our model is reset to an EMA but the EMA is never reset. We also compare to the simplest reset idea, \texttt{Reset to Init}, and reset our policy to the initial model. We run all methods as previously and plot the pareto graph in Figure~\ref{fig:imdb-ablation-reset}, for task and drift graphs see Appendix~\ref{app:ablations}. We find that even simple resets are already performant but the two ablations have a tradeoff: \texttt{Reset to EMA} is better at lower reward because it maintains performance whereas \texttt{Reset to Init} does better at higher reward because it doesn't drift as much. Elastic Reset combines the benefits of both and outperforms each method.

Next, we consider our method's robustness to hyperparameters. First, we search along different EMA decay rates $\eta$ and plot our results in Figure~\ref{fig:imdb-ablation-decay} finding that our method is quite robust to choice of decay. Next, we investigate robustness to the choice of KL penalty coefficient $\beta$. We search across coefficients that range from 10x smaller to 10x larger than our best KL penalty coefficient for PPO ($\beta = 0.1$) and Elastic Reset ($\beta = 0.001$). For visual clarity, we plot PPO in Figure~\ref{fig:imdb-klablation-ppo} and Elastic Reset in Figure~\ref{fig:imdb-klablation-emaset} and only plot four points for PPO $\beta = 0, 0.01$ to maintain visual scale. We find that PPO is not robust to choice of KL and larger values correspond to better pareto curves but slower training. Results with NLPO are similar and shown in Appendix~\ref{app:ablations}. In contrast, Elastic Reset seems to be more robust to choice of KL with $0.001$ producing the best curves while 10x larger and smaller values are similar. As opposed to PPO, Elastic Reset even works reasonably well without a KL penalty at all, ($\beta = 0$), matching PPO's best performance with a KL. This demonstrates that the expensive KL penalty may be replaced with the cheap Elastic Reset, although the combination of the two is best. This is also in line with previous work that have argued that the KL penalty may be unnecessary for RLHF \citep{bai_training_2022,gao_scaling_2022}. We also ablate the frequency of resets in Appendix~\ref{app:reset-freq} and find that pareto curves are essentially unchanged. 

Finally, we provide an empirical intuition for Elastic Reset: in Appendix~\ref{app:explain-value} we show that resets iteratively improve the value function and in Appendix~\ref{app:explain-ema} we show how EMA smoothes optimization but requires resetting in order to achieve high performance.

\begin{figure}
     \centering
      \begin{subfigure}[b]{0.32\linewidth}
         \centering
         \includegraphics[width=\linewidth]{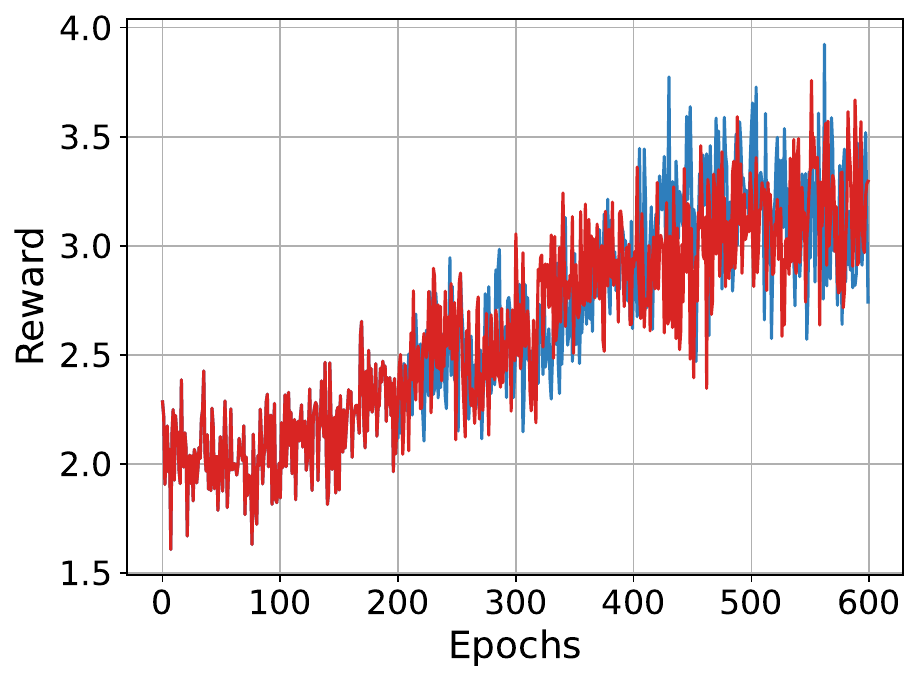}
         \caption{Reward}
         \label{fig:stackllama-rew}
     \end{subfigure}
      \begin{subfigure}[b]{0.32\linewidth}
         \centering
         \includegraphics[width=\linewidth]{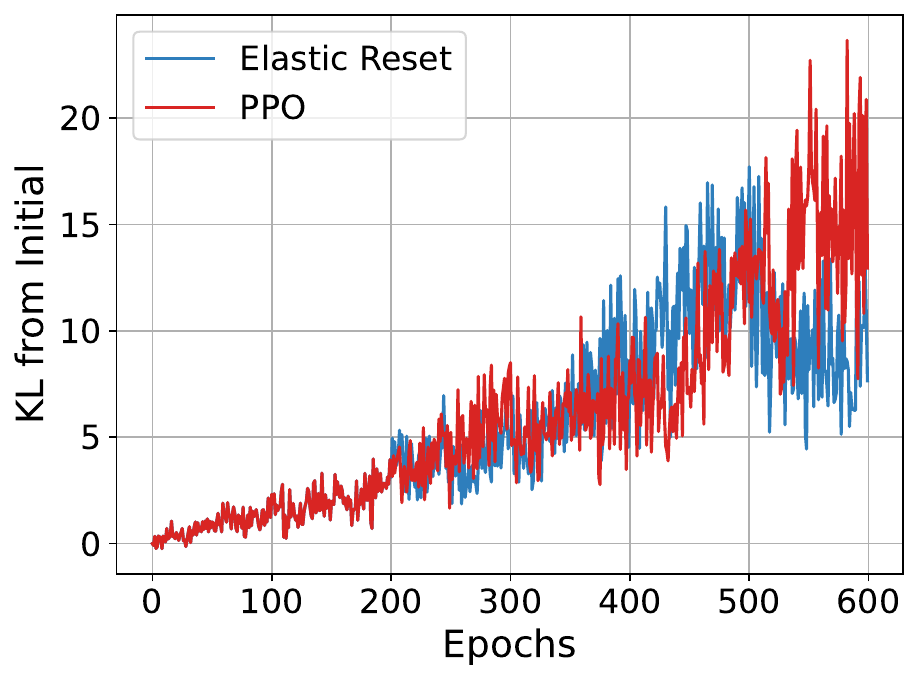}
         \caption{Drift}
         \label{fig:stackllama-kl}
     \end{subfigure}
     \begin{subfigure}[b]{0.32\linewidth}
         \centering
         \includegraphics[width=\linewidth]{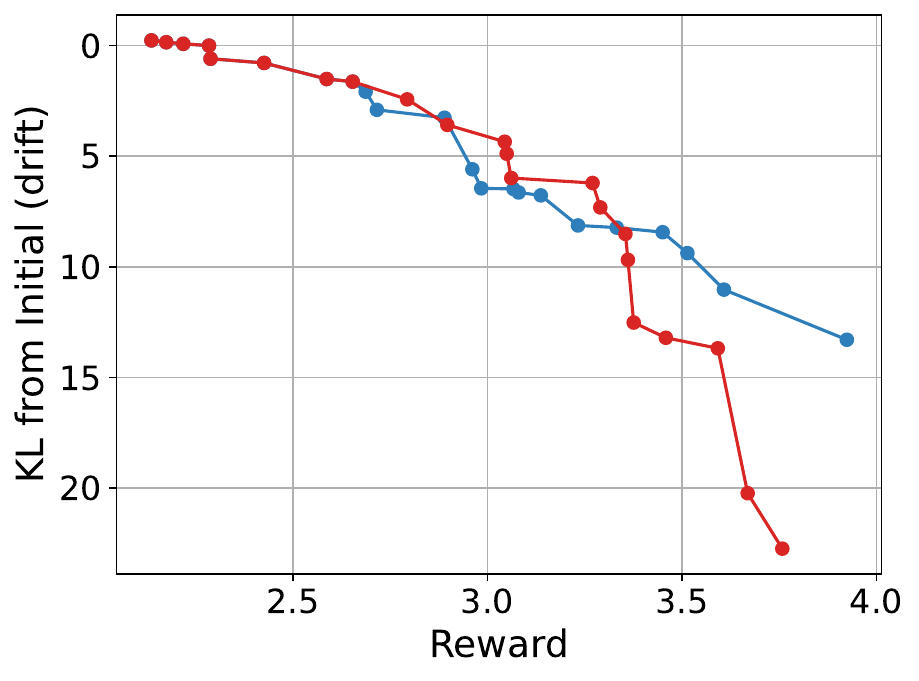}
         \caption{Pareto Frontier Graph}
         \label{fig:stackllama-pareto}
     \end{subfigure}
    \caption{Elastic Reset compared to PPO on StackLLaMA: A LLaMA-7B model RLHF finetuned on StackExchange as a helpful, technical QA chatbot}
    \label{fig:stackllama}
\end{figure}

\section{StackLLaMA: Practical RLHF}

\paragraph{Setup} Finally, we apply Elastic Reset to a larger-scale RLHF pipeline. We choose LLaMA \citep{touvron_llama_2023} as it is a prominent open-source model that has demonstrated strong performance on benchmarks. We follow \citet{beeching_stackllama_2023} to finetune LLaMA-7B with RLHF on the StackExchange dataset \citep{lambert_huggingface_2023} to output helpful answers to technical questions, as judged by humans. Users ask technical questions on StackExchange and upvote the best answers. We score answers from StackExchange based on the number of upvotes they received from users, \texttt{ score }$= \log_2(1 + \texttt{upvotes})$ \citep{askell_general_2021} . At most 10 answers are drawn per question, text is cleaned, and HTML is converted to Markdown to make it easier to parse. First, we finetune LLaMA-7B with language modelling on the dataset to get LLaMA-7B-SE. We then further finetune it to get a reward model by learning to predict which of two answers was more upvoted \citep{stiennon_learning_2020}. For a given question $x$ and two answers $y_{+,-}$ (where $y_+$ is preferred), the loss for our reward model $r_\theta$ is $\log \sigma (r_\theta(x,y_+) - r_\theta(x,y_-))$.  Finally, we finetune LLaMA-7B-SE with RL against the reward model by sampling questions from the dataset and learning to optimize the reward for our model's answer. All finetuning is done with a converted 8-bit model \citep{dettmers_llmint8_2022} and LoRA \citep{hu_lora_2021} for efficiency and to make the model training fit on our GPUs. We rely on the HuggingFace trl \citep{von_werra_trl_2023} and peft \citep{mangrulka_peft_2023} libraries. All technical details are described in Appendix~\ref{app:stackllama}.

\paragraph{Experiment} We again compare to PPO with \citet{ziegler_fine-tuning_2019} modifications i.e. KL penalty with a dynamically decaying coefficient. We run for 600 epochs (equivalent to 100k updates) and Elastic Reset every 260 epochs to get two resets / three iterations. Each run takes 20 hours on 4 A100s. Since only the LoRA parameters are being learned, we use Elastic Reset on those and therefore maintain only a small percentage of parameters in our EMA. We use a decay rate $\eta = 0.995$ and a KL penalty coefficient $\beta = 0.02$ for both methods. Calculating perplexity for each epoch is computationally infeasible so we measure drift during training with the KL from the pretrained model over samples as done previously in other larger-scale RLHF \citep{bai_training_2022,gao_scaling_2022}. 

\paragraph{Results} We plot reward in Figure~\ref{fig:stackllama-rew} \footnote{Note that our reward results differ from the original StackLLaMA likely due to a pending issue in their code affecting their reward model. After fixing the issue, the only difference seems to be an increase in base reward, and our reward curves are visually quite similar to theirs. See Appendix~\ref{app:stackllama} for details} and KL from initial model over training in Figure~\ref{fig:stackllama-kl}. As noted by \citet{beeching_stackllama_2023}, the task is much noisier at a larger scale and with a real HF reward model. As a sanity check, we find that Elastic Reset tracks PPO until the first reset at 260 epochs where it drops only slightly, but also doesn't lose much performance. Around the second reset at 520 epochs, we see a much sharper drop but also maintaining the same approximate reward. At the end, Elastic Reset provides a non-trivial reduction in drift while aligning just as well as PPO. The pareto curve in Figure~\ref{fig:stackllama-pareto} shows Elastic Reset is equal or slightly worse at low reward but shows large improvements over PPO at higher reward. Notably, Elastic Reset seems to work out-of-the-box with LoRA. To evaluate drift another way, we get the perplexity of the final models over the StackExchange validation set as in Section~\ref{sec:imdb}. For a more robust view of reward, we train two more reward models using different seeds and evaluate the increase in reward between initial and final models. We show mean and standard deviation across the three reward models in Table~\ref{tab:stackllama}, we find that Elastic Reset achieves a slightly better final reward than PPO while maintaining lower perplexity on the data. To examine a true alignment tax, we run our models on HumanEval \cite{chen_evaluating_2021}, a programming benchmark that provides another view of drift. Answering human-written coding questions is both a useful capability for our model and also falls within a similar domain to StackExchange. The benchmark tests for functional correctness such that pass@1 corresponds to the percentage of problems solved by the model on the first try as shown in Table~\ref{tab:stackllama}. Training with PPO degrades performance compared to the initial model, demonstrating a large alignment tax. In contrast, Elastic Reset achieves a similar reward but maintains performance, even slightly improving on pass@10, creating an alignment bonus \citep{askell_general_2021} instead of tax. 

\begin{table}[t]
\caption{Evaluations of the initial (zero-shot) and finetuned StackLLaMA models after 600 epochs. We measure alignment using an average over three reward model trained with three different seeds and drift with perplexity on the data. HumanEval is a programming benchmark that acts as a practical measure of drift / alignment tax. }
\label{tab:stackllama}
\begin{center}
\begin{sc}
% \resizebox{0.5\linewidth}{!}{
\begin{tabular}{lccc}
\toprule
 & $\uparrow \Delta$ Reward & $\downarrow$ Perplexity & $\uparrow$ HumanEval (pass@1,pass@10) \\
\midrule
Zero-shot & 0 & 4.43 & 11.0, 12.7   \\
\midrule
PPO  &  0.81 $\pm$ 0.06 & 4.62 &  ~~7.8, 10.7  \\
\textbf{Elastic Reset} & 0.96 $\pm$ 0.09  & 4.57  &  11.0, 13.0 \\
% PPO + KL EMA     & .620$\pm$0.05 & 33.73$\pm$0.45 \\
% + \textbf{Reset to EMA}      & \textbf{.636$\pm$0.03} & \textbf{33.65$\pm$0.31} \\
% + \textbf{Elastic Reset}     & \textbf{.611$\pm$0.02} & \textbf{33.32$\pm$0.23} \\ 
\bottomrule 
\end{tabular}
% }
\end{sc}
% \scriptsize	
%\end{small}
\end{center}
\vskip -0.1in
\end{table}

\section{Limitations}

As a method, Elastic Reset is quite cheap computationally because both EMA updates and resets take negligable time compared to RLHF training and the EMA model can be stored on CPU. But our method is sensitive to the choice of reset rate; we chose heuristically based on when it seemed the model was overfitting. It is also possible to reset the policy and EMA model at different time scales, which could be a source of improvement. Our method also resets all of the trainable parameters, research in similar methods suggests that resetting larger models can benefit from resetting only part of the network \citep{zhou_fortuitous_2022,nikishin_primacy_2022} or weighted-averaging instead of resets \citep{doro_sample-efficient_2023}. We leave both of these directions to future work.

Although we have thoroughly investigated our method on three different tasks, we note that none of them are ideal RLHF benchmarks. As pointed out by \citet{gao_scaling_2022}, we measure our model's performance using the same reward model we optimize. This can lead to reward model overoptimization and our metric could mask overfitting and lack of generalization to the real world i.e. actual human preferences. An ideal benchmark could include a ``gold'' reward model as a proxy for human preference \citep{gao_scaling_2022}, but no such benchmarks are open-sourced and available. 

Finally, we note that we follow all previous RLHF work and investigate only on-policy methods \citep{ziegler_fine-tuning_2019, stiennon_learning_2020, askell_general_2021, bai_training_2022}. Previous work in resetting for RL has focused on off-policy methods and demonstrated strong performance \citep{nikishin_primacy_2022,doro_sample-efficient_2023}. As previously noted, our method can be seen as an adaptation of those to on-policy RL. In RLHF, PPO is by far the most popular method and on-policy is the dominant paradigm since it guarantees better local gradients. But it is possible that off-policy methods could implicitly balance performance and drift by incorporating a replay buffer with older data.

\section{Conclusion}

The problems of drift \citep{lee_countering_2019}, alignment tax \citep{askell_general_2021}, reward model overoptimization \citep{gao_scaling_2022}, and reward hacking \citep{clark_faulty_2016} are inherent to RLHF and reduce its efficacy. We have introduced a simple but powerful new method, Elastic Reset, to tackle this problem and improve performance while maintaining linguistic capabilities. We have shown its ability on three different tasks and across three different scales: from 6 layer Transformers to GPT2 to LLaMA-7B. The problem of drift is currently being addressed with a standard KL penalty despite the computational cost, tradeoff with reward, and recent claims that is may be unnecessary \citep{bai_training_2022,gao_scaling_2022}. 
% A larger coefficient for the KL penalty also slows down training, increasing the time to achieve the same reward \citep{gao_scaling_2022}. 
% Yet KL penalties remain the dominant paradigm in RLHF. 
Elastic Reset is a cheap and effective method to tackle the same problem, achieving a better tradeoff of reward and drift while reducing alignment tax. We hope our method leads to better RLHF and therefore models that are closer aligned with human preferences \citep{ziegler_fine-tuning_2019}. As well, we hope this work invigorates more research into improving the reward / drift tradeoff of RLHF with a focus on computationally efficient methods that scale.

% The simplicity of Elastic Reset allows it to be relatively easily applied to other, similar problems. One perspective on drift is that of catastrophic forgetting \citep{robins_catastrophic_1995} where the model learns the reward but forgets previous knowledge. Therefore, Elastic Reset may be effective in domains such as continual learning, where neural networks must learn new tasks while maintaining performance on older tasks \citep{ring_child_1997}. 

\begin{ack}
MN is supported by Fonds de recherche du Québec – Nature et technologies and Sony. MN would like to thank Issam Laradji, ServiceNow Research, Mila, and Compute Canada for providing resources used in the experiments.
\end{ack}

% \section{Supplementary Material}

% Authors may wish to optionally include extra information (complete proofs, additional experiments and plots) in the appendix. All such materials should be part of the supplemental material (submitted separately) and should NOT be included in the main submission.

% \section*{References}

\bibliography{references}

\clearpage
\appendix
\section{Translation Game}
\label{app:translation-game}

\subsection{Experimental Details}
\label{app:translation-game-details}
We implement the translation game in the fairseq library \citep{ott_fairseq_2019} on top of Pytorch \citep{paszke_pytorch_2019}. All experiments are run with 5 seeds where each run uses a single 16G V100 GPU. All plots show the mean and standard deviation over seeds. We score BLEU using detokenized sacreBLEU \citep{post_call_2018}. Plots are made with Matplotlib \citep{hunter_matplotlib_2007}

\subsection{Translation Game with Transformers}
\label{app:translation-game-transformers}

Note that \citet{lee_countering_2019} (as well as previous work on the Translation Game) use seq2seq \citep{sutskever_sequence_2014} LSTMs \citep{hochreiter_long_1997} with attention \citep{bahdanau_neural_2015}. For consistency with other experiments and relevance to mainstream research we switch to a Transformer architecture. LSTMs results are similar and results are available in Appendix \ref{app:translation-game-lstm}. Also note that we use REINFORCE as our estimator but Gumbel-Softmax \citep{jang_categorical_2017,maddison_concrete_2017} is also feasible \citep{lu_countering_2020} and preliminary results have been similar.

We follow \citet{lee_countering_2019} for preprocessing both IWLST and Multi30k. We use the \texttt{iwslt-de-en} architecture from the fairseq library \citep{ott_fairseq_2019} with their default IWSLT DE-EN hyperparameters and training, specifically training with AdamW \citep{loshchilov_decoupled_2022} using default hyperparameters. We pretrain our models on IWSLT following \citet{lee_countering_2019} and early-stop on the tst2013 validation set. Our final validation scores are 43.85 BLEU for FR$\to$EN and 29.4 BLEU for EN$\to$DE. 

\begin{figure}
     \centering
      \begin{subfigure}[b]{.32\linewidth}
         \centering
         \includegraphics[width=\linewidth]{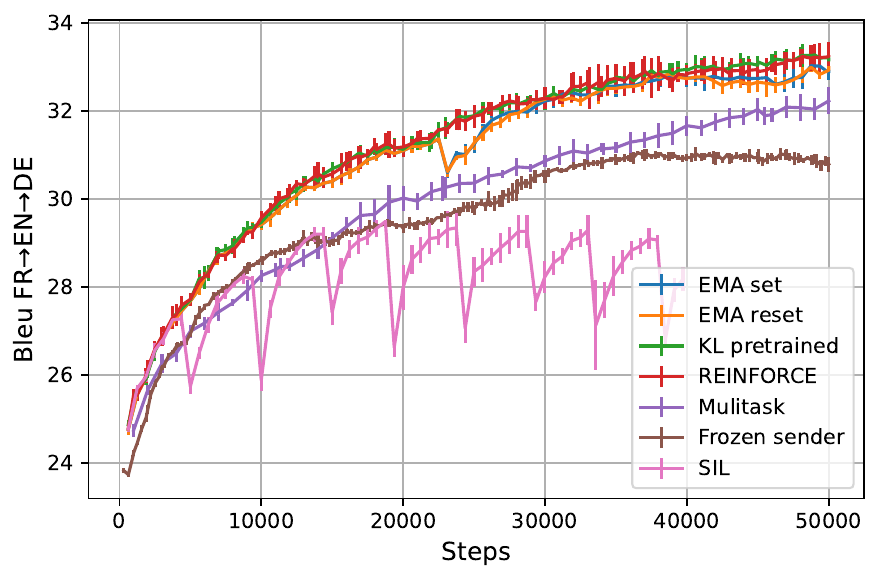}
         \caption{Task Performance}
     \end{subfigure}
     \begin{subfigure}[b]{.32\linewidth}
         \centering
         \includegraphics[width=\linewidth]{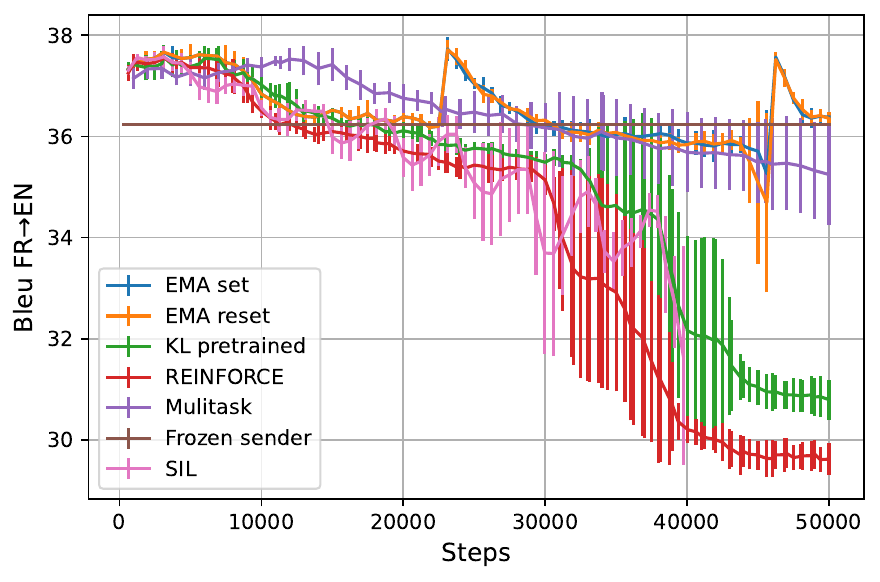}
         \caption{Language Drift}
     \end{subfigure}
     \begin{subfigure}[b]{.32\linewidth}
         \centering
         \includegraphics[width=\linewidth]{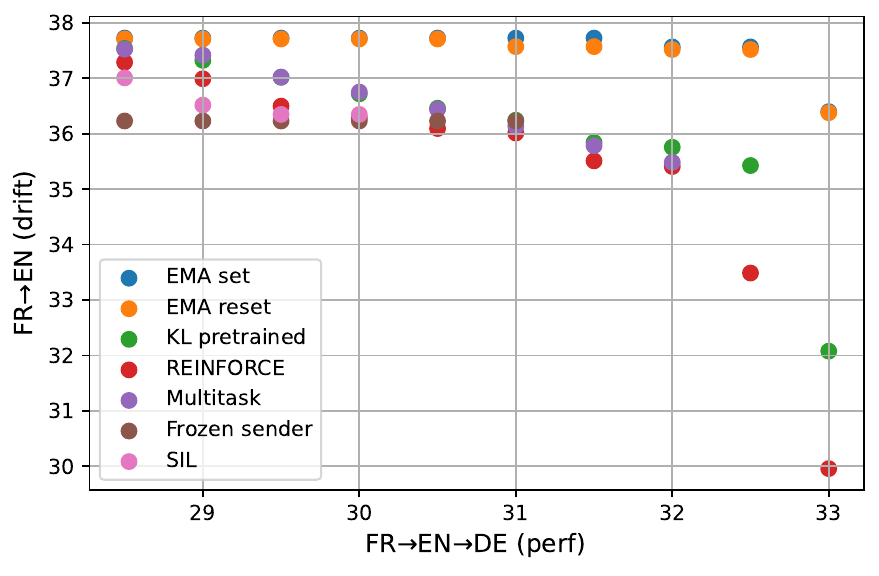}
         \caption{Pareto Frontier Graph}
     \end{subfigure}
    \caption{All transformer model methods on the Translation Game. We measure (a) Task Performance with FR$\to$EN$\to$DE BLEU and (b) Language Drift with FR$\to$EN BLEU, on the validation set during finetuning. We plot the mean and show error bars for standard deviation over 5 seeds. To compare how methods do on both metrics, we plot (c) the best achieved drift vs task performance across finetuning.}
    \label{fig:translation-game-transformers-appendix}
\end{figure}

We report all final validation scores in Table \ref{tab:translation-game-transformers} and all curves in Figure \ref{fig:translation-game-transformers-appendix}.

Given that we have a learning reward model (EN$\to$DE), iteratively resetting our FR$\to$EN model to its pretrained weights can be an effective strategy \citep{rita_emergent_2022}. We report this as \textsc{Reset} and find it performs quite well but not as well as Elastic Reset. We did not include these results in the main paper because we focused on methods that could also work with fixed reward models.

\begin{table}[h]
\caption{Translation Game final validation scores }
\label{tab:translation-game-transformers}
% \vskip -0.2in
\begin{center}
\begin{small}
\begin{sc}
\begin{tabular}{lcc}
\toprule
 & $\uparrow$ FR$\to$EN$\to$DE & $\uparrow$ FR$\to$EN \\
\midrule
pretrained      & 23.8 & 36.2 \\
frozen sender   & 30.8$\pm$0.2 & \textbf{36.3$\pm$0.1}  \\
REINFORCE       & \textbf{33.2$\pm$0.3} & 29.6$\pm$0.3 \\
+ SIL           & 28.2$\pm$0.4 & 27.3$\pm$4.4 \\
+ multitask (S2P) & 32.2$\pm$0.3  & 35.2$\pm$1.0  \\
+ KL pretrained & \textbf{33.2$\pm$0.2}  & 30.8$\pm$0.4  \\
+ Reset         & 32.5$\pm$0.1   & 33.3$\pm$0.1  \\
+ Reset to EMA  & 32.9$\pm$0.1 & \textbf{36.3$\pm$0.1} \\
+ Elastic Reset & 33.0$\pm$0.1 & \textbf{36.3$\pm$0.1} \\
\bottomrule
\end{tabular}
\end{sc}
\end{small}
\end{center}
\vskip -0.25in
\end{table}

\subsection{Translation Game with LSTMs}
\label{app:translation-game-lstm}

We follow \citet{lee_countering_2019} for preprocessing and pretraining on IWSLT. We compare our pretraining results to both \citet{lee_countering_2019} and \citet{lu_countering_2020} in Table \ref{tab:iwslt-lstm}

\begin{table}[h]
\caption{BLEU score of IWSLT-pretrained LSTM models on IWSLT 2013 validation set}
\label{tab:iwslt-lstm}
% \vskip -0.2in
\begin{center}
\begin{small}
\begin{sc}
\begin{tabular}{ccc}
\toprule
& FR$\to$EN & EN$\to$DE  \\
\hline 
\citet{lee_countering_2019} & 34.1 & 22.0 \\
\citet{lu_countering_2020} & 32.2 & 20.2 \\
Ours & 38.5 & 23.2 \\
\bottomrule
\end{tabular}
\end{sc}
\end{small}
\end{center}
\vskip -0.25in
\end{table}

Next we finetune on Multi30k using the same hyperparameters as \citet{lu_countering_2020}. We plot results in Table \ref{tab:multi30k} and compare to published numbers from previous work. As usual, drift is the negative change in FR$\to$EN BLEU from the pretrained model. Task performance is the positive change in FR$\to$EN$\to$DE BLEU from the pretrained models. Combined is the sum of drift and task performance. We show the pretrained, baseline, and best-performing model from previous work. Note that our results are not directly comparable to previous work because we evaluate using detokenized sacreBLEU \citep{post_call_2018} whereas previous work wrote their own BLEU evaluation code and did not detokenize.

\begin{table}[h]
\caption{BLEU scores and $\pm$ standard deviation on the Multi30k Translation Game using IWSLT-pretrained LSTM models.}
\label{tab:multi30k}
% \vskip -0.2in
\begin{center}
\begin{small}
\begin{sc}
\begin{tabular}{llccccc}
\toprule
\centering
     & Method & FR$\to$EN & FR$\to$EN$\to$DE & Drift & Perf & Combined \\
    \hline 
    \citet{lee_countering_2019} & pretrained & 27.2 & 16.3 & & &  \\
     & REINFORCE & $12.4 \pm 0.7 $ & $24.5 \pm 1.5 $ & -14.8 & +8.2 & -6.6 \\
     & + LM & $23.6 \pm 1.1 $  & $27.7 \pm 0.4 $ & -3.6 & +11.4 & +7.8 \\
     & + LM + G & $24.8 \pm 0.4 $ & $28.1 \pm 0.7 $ & -2.4 & +11.8 & +9.4 \\
     \hline
    \citet{lu_countering_2020} & pretrained & 29.4  & 15.7 & & &  \\
    & gumbel-softmax & $14.5 \pm 0.8 $ & $27.1 \pm 0.1 $ & -14.9 & +11.4 & -3.5  \\ 
    & SIL & $29.4 \pm 0.3 $ & $28.3 \pm 0.2 $ & 0 & +12.6 & +12.6 \\
    \hline
    \textbf{Ours} & pretrained & 32.6 & 18.0 & & & \\
    & frozen-sender & $32.6 \pm 0 $  & $25.1 \pm 0.1 $ & 0 & +7.1 & +7.1  \\
    & REINFORCE & $30.0 \pm 0.2 $  & $30.3 \pm 0.2 $ & -2.6 & +12.3 & +9.7 \\
    & LM $\lambda = 0.01$ & $31.5 \pm 0.1 $ & $30.2 \pm 0.2 $ & -1.1 & +12.2 & +11.1 \\
    & Multitask $\lambda = 0.1$ & $32.9 \pm 0.2 $ & $28.5 \pm 0.3 $ & +0.3 & +10.5 & +10.8 \\
    & SIL & $27.7 \pm 0.7 $  & $24.3 \pm 0.1 $ & -4.9 & 6.3 & +1.4 \\
    & Elastic Reset & $32.6 \pm 0.1 $  & $30.0 \pm 0.1 $ & 0 & +12.0 & \textbf{+12.0} \\
    \hline
\end{tabular}
\end{sc}
\end{small}
\end{center}
\vskip -0.25in
\end{table}

Our results for the baseline are notably better than \citet{lee_countering_2019,lu_countering_2020}. The only difference between our code and theirs as far as we can tell is (1) we use an exponential moving average baseline for REINFORCE whereas \citep{lee_countering_2019} use an Actor-Critic method, (2) \citet{lu_countering_2020} uses 0.1 gradient clipping and we do not use gradient clipping
(3) in preprocessing Multi30k, we first tokenize \citep{koehn_moses_2007} then lowercase whereas previous works did the opposite order.

A more reasonable explanation for the improvement in results is how we choose to evaluate. Previous work simply ran all methods for the same number of updates but this doesn't account for, even implicit, hyperparameter optimization. Previous methods show that the baseline's FR$\to$EN$\to$DE BLEU scores plateau and there is significant language drift in FR$\to$EN without real improvements to the task score. We hypothesize that these extra training episodes only serve to increase the drift without measuring what we actually care about: performance gain for drift. Since the number of updates is arbitrary, we believe that early stopping on a reasonable metric is a better evaluation protocol and choose hyperparameters such that methods plateau at the end.

We also note that our results with the SIL method of \citet{lu_countering_2020} are negative. We do not manage to gain any improvement in performance. We collaborated with the authors of \citet{lu_countering_2020} for many months but, in our setup using the fairseq library \citep{ott_fairseq_2019} could not reproduce their results. One of their fundamental results is that a student sender can outperform a teacher sender that it is distilling from. We could not reproduce this and believe it is a difference in the pretrained models.

\section{IMDB Mock Sentiment}
\label{app:imdb}

\begin{figure}[h]
     \centering
     \begin{subfigure}[b]{0.32\linewidth}
         \centering
         \includegraphics[width=\linewidth]{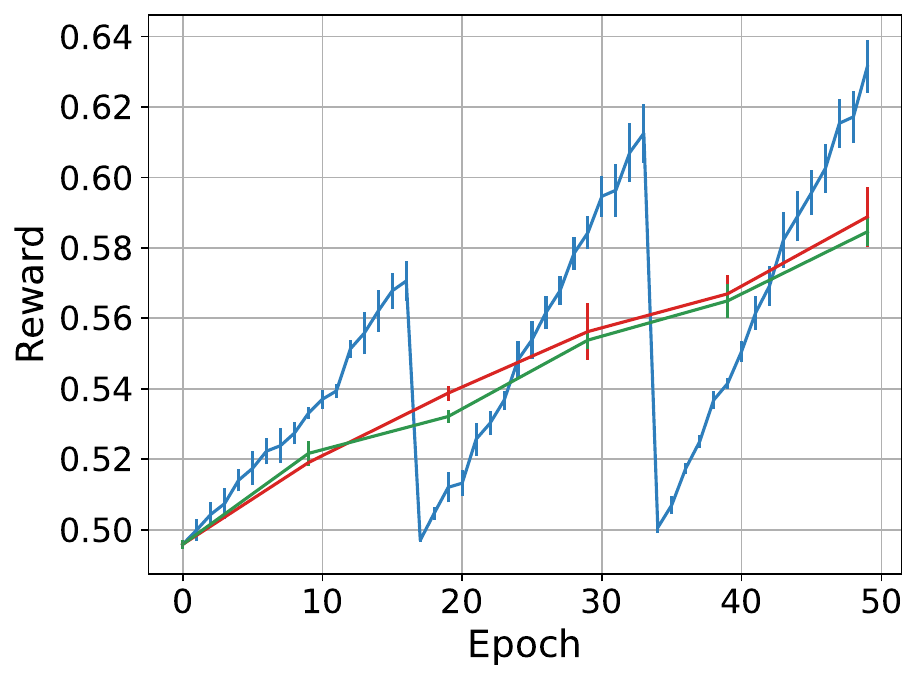}
         \caption{Task Performance}
         \label{fig:imdb-sentiment}
     \end{subfigure}
     \begin{subfigure}[b]{0.32\linewidth}
         \centering
         \includegraphics[width=\linewidth]{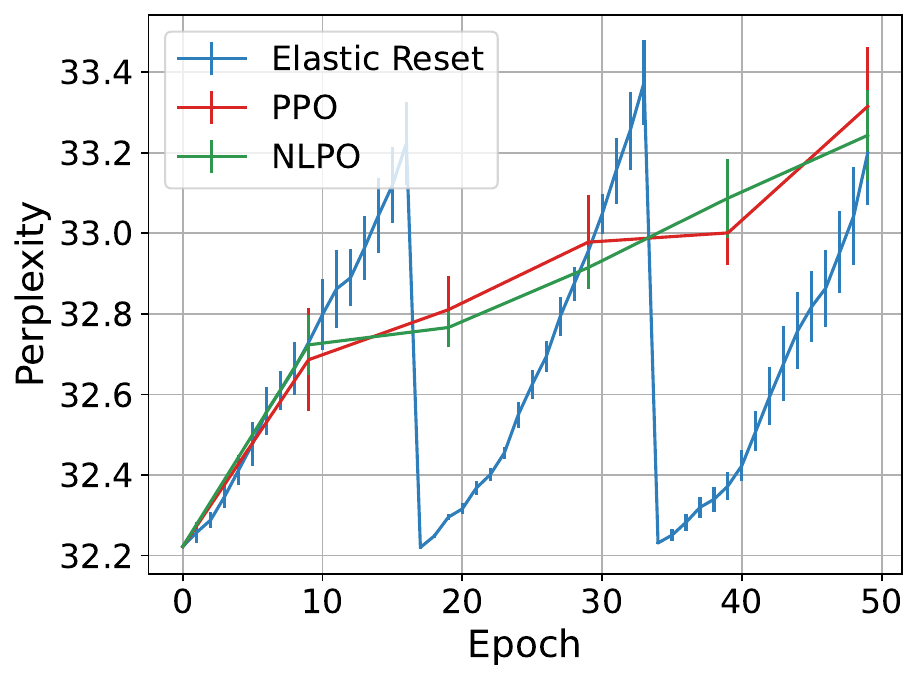}
         \caption{Language Drift}
         \label{fig:imdb-ppl}
     \end{subfigure}
     \begin{subfigure}[b]{0.32\linewidth}
         \centering
         \includegraphics[width=\linewidth]{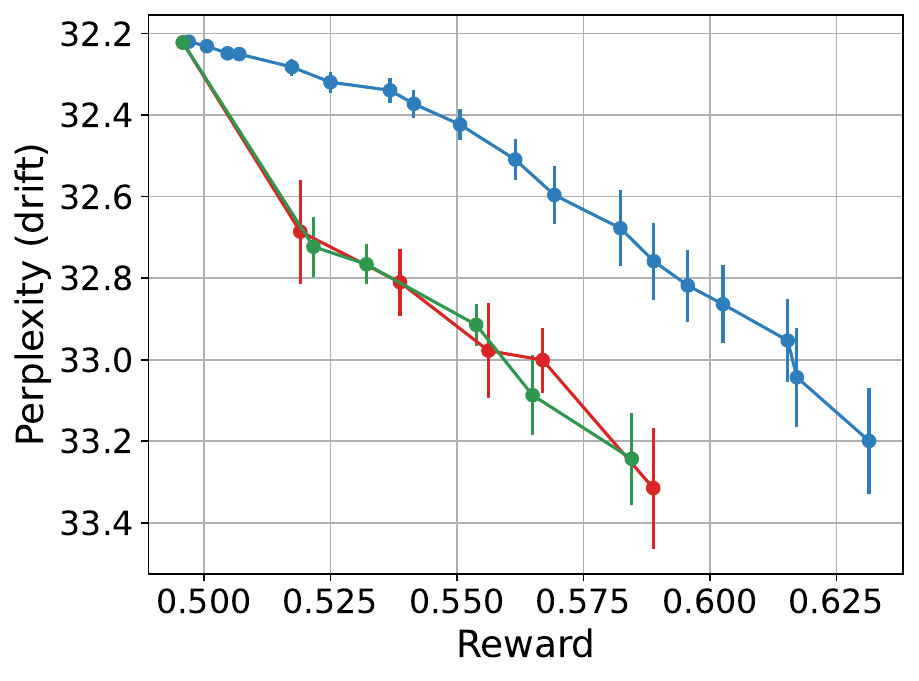}
         \caption{Pareto Frontier Graph}
         \label{fig:imdb-pareto}
     \end{subfigure}
    \caption{IMDB mock sentiment validation set scores for Elastic Reset, PPO, and NLPO. We measure (a) Language Drift and (b) Task Performance via Semantic Score on the validation set over finetuning. All methods also include a KL penalty. We plot mean and standard error across 5 seeds.}
    \label{fig:imdb-nlpo}
    \vskip -0.1in
\end{figure}

\subsection{Experimental Details}

We run experiments and implement our method in the RL4LMs library \citep{ramamurthy_is_2022}. All experiments are run with 5 seeds where run uses a single 40G A100 GPU. All plots show the mean and standard error over seeds. 

For PPO, we use the default hyperparemters provided by \citet{ramamurthy_is_2022}. We also compared to NLPO, we found the defaults had a mistake and after communication with \citet{ramamurthy_is_2022} we changed the learning rate to 1e-6 and target update iterations to 50. We plot results in Figure~\ref{fig:imdb-nlpo}. This still did not manage to reproduce the original NLPO test scores from \citet{ramamurthy_is_2022} but we found that our validation curves matched their provided curves in the appendix. After communications \citep{ammanabrolu_re_2023}, we both agreed that we should use our reproducible test scores for NLPO in lieu of the original test scores.

Our best Elastic Reset hyperparameters are the default PPO parameters (\verb|gpt2_ppo.yml|) from RL4LMs \citep{ramamurthy_is_2022} with a few modifications: target kl is set from 0.5 to 1.0 and KL coefficient is set much lower to 0.001. We use an EMA decay of 0.995 and reset every 17 epochs. Configs to reproduce the experiments can be found in the RL4LMs folder under \texttt{scripts/training/task\_configs/imdb\_text\_continuation}

\section{StackLlama}
\label{app:stackllama}

\begin{figure}[h]
     \centering
      \begin{subfigure}[b]{.4\linewidth}
         \centering
         \includegraphics[width=\linewidth]{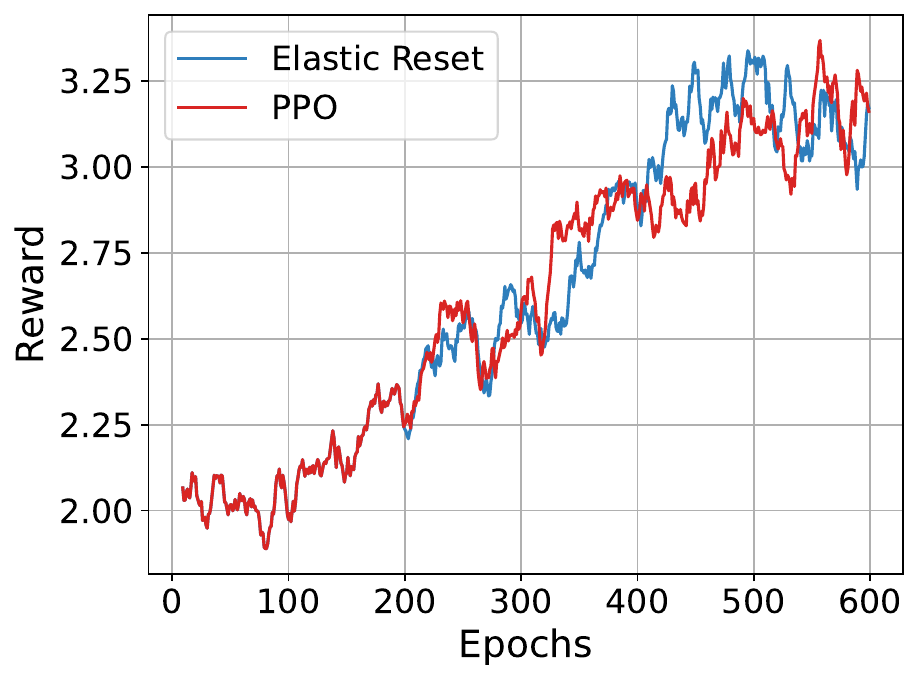}
         \caption{Reward}
     \end{subfigure}
     \begin{subfigure}[b]{.4\linewidth}
         \centering
         \includegraphics[width=\linewidth]{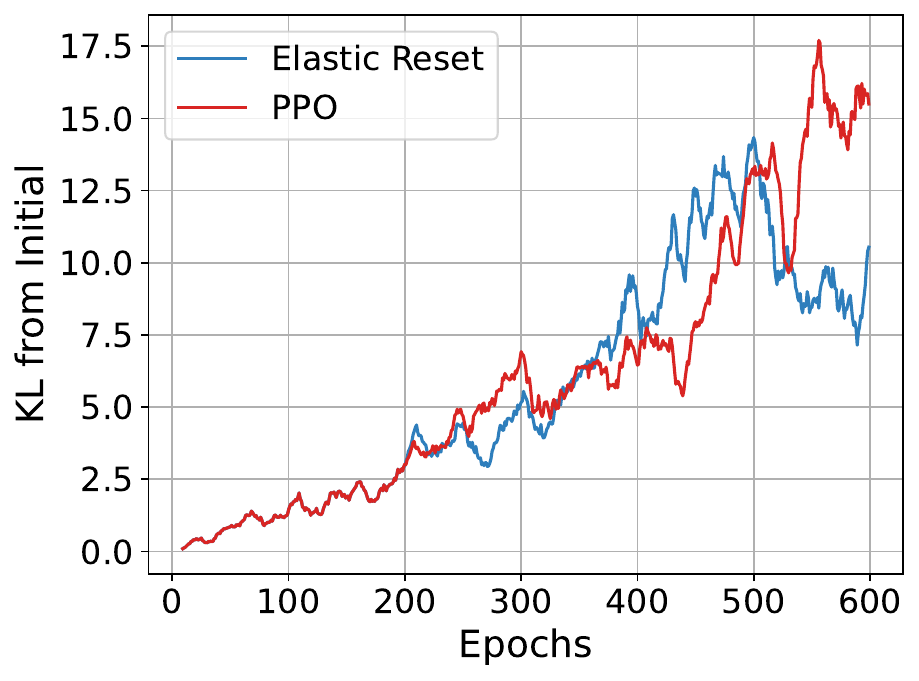}
         \caption{KL from Initial}
     \end{subfigure}
    \caption{Rolling Window Mean (window = 10) version of the StackLLaMA results}
    \label{fig:stackllama-rollingmean}
\end{figure}

\subsection{Experimental Details}

We follow the original StackLlama and use the trl library from Huggingface on top of the transformers library \citep{wolf_transformers_2020} to train the model on the StackExchange dataset, loaded with datasets \citep{lhoest_datasets_2021}. We use the original authors' LoRA adapter weights to create our supervised model LLaMA-7B-SE but train our own reward model as there were issues loading the pretrained reward model. As noted by \citep{beeching_stackllama_2023}, the reward modelling task is difficult enough that humans struggle with it and our final model achieves 64\% accuracy compared to the original 67\%. Although we follow the original authors method and use their codebase, we note that our results may be different but valid. There have been many updates and fixes to the \texttt{trl} codebase since the authors' original blog post and specifically a possible issue in the code for creating reward models could have affected the original authors' run.

Furthermore, to speed up training, we used a 2x smaller KL coeffient and ran with half the number of GPUs. Specifically, RL training is run on 4x 80G A100 GPUs for 20 hours using Accelerate \citep{gugger_accelerate_2022}. We evaluate perplexity of the model on the validation set used in supervised finetuning, 4000 examples from the supervised training dataset. The configs for reproducing our experiments are in the \texttt{trl} library folder under \texttt{examples/stackllama/scripts/configs}

For visual clarity, we provide a version of our StackLLaMA results with a rolling window mean in Figure~\ref{fig:stackllama-rollingmean}

\subsection{HumanEval}

Our measure of alignment tax, HumanEval \citep{chen_evaluating_2021}, is a programming benchmark where each question was hand-written by humans (OpenAI engineers) to be unseen in training data. We believe these questions were unseen in LLaMA training data as well. We prompt the model with the question and it writes the corresponding code. The measure of success is functionally correct code i.e. we actually execute the code LLaMA wrote and see if it gets the right output. Since we decode by sampling, we generate $N = 100$ continuations of the prompt and then follow \citet{chen_evaluating_2021} to estimate how often our model would get the right answer on the first generated continuation (pass @ 1) and within the first 10 generated continuations (pass @ 10). We evaluate HumanEval using CodeCapybara's evaluation harness \citep{to_codecapybara_2023}.

\section{Ablations Full Graphs}

\subsection{Reset Number}
\begin{figure}[ht]
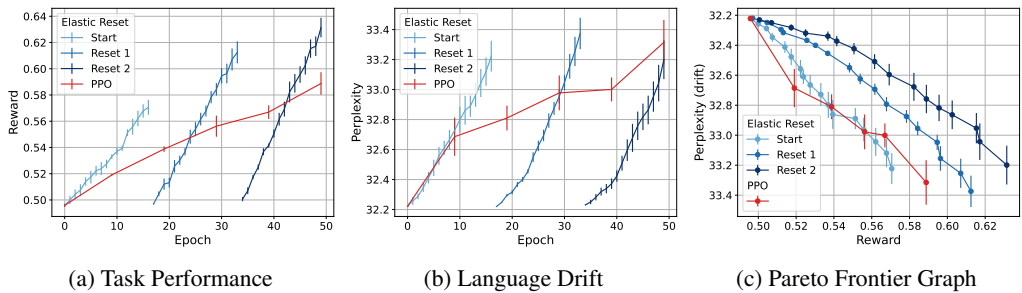

     \centering
      \begin{subfigure}[b]{.32\linewidth}
         \centering
         \includegraphics[width=\linewidth]{images/imdb_compute_task.pdf}
         \caption{Task Performance}
     \end{subfigure}
     \begin{subfigure}[b]{.32\linewidth}
         \centering
         \includegraphics[width=\linewidth]{images/imdb_compute_drift.pdf}
         \caption{Language Drift}
     \end{subfigure}
     \begin{subfigure}[b]{.32\linewidth}
         \centering
         \includegraphics[width=\linewidth]{images/imdb_compute_pareto.pdf}
         \caption{Pareto Frontier Graph}
     \end{subfigure}
    \caption{Plotting PPO vs Elastic Reset on IMDB but splitting the results visually between resets.  Users with less compute can choose fewer resets. We didn't find improved results after 2 resets for IMDB}
    \label{fig:imdb-resetsplit-emaset}
\end{figure}

\subsection{Reset Type}
\begin{figure}[H]
     \centering
      \begin{subfigure}[b]{.32\linewidth}
         \centering
         \includegraphics[width=\linewidth]{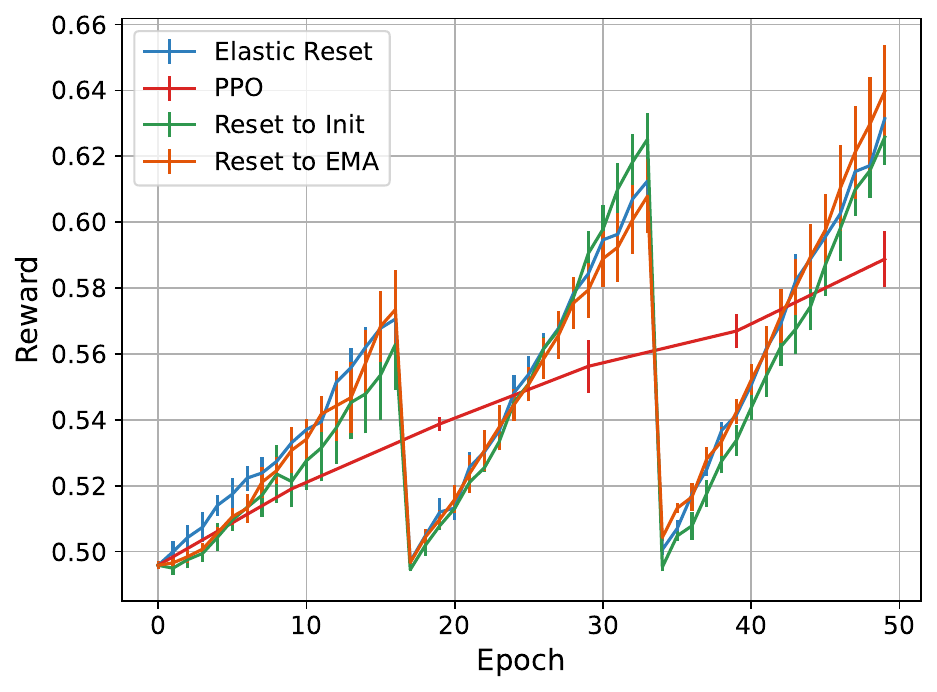}
         \caption{Task Performance}
     \end{subfigure}
     \begin{subfigure}[b]{.32\linewidth}
         \centering
         \includegraphics[width=\linewidth]{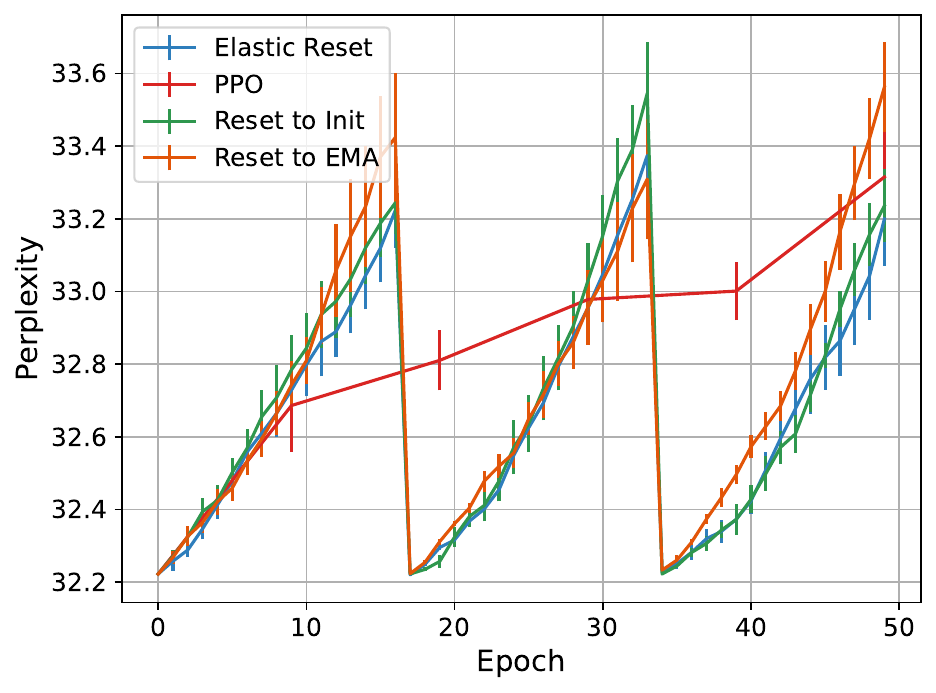}
         \caption{Language Drift}
     \end{subfigure}
     \begin{subfigure}[b]{.32\linewidth}
         \centering
         \includegraphics[width=\linewidth]{images/imdb_ablation_reset_pareto.pdf}
         \caption{Pareto Frontier Graph}
     \end{subfigure}
    \caption{Ablation of Elastic Reset, Reset to EMA, and Reset to Init on IMDB. We plot the mean and show error bars for standard error over 5 seeds.}
    \label{fig:imdb-resetablation}
\end{figure}

\subsection{KL Coefficient}
\label{app:ablations}

\begin{figure}[ht]
     \centering
      \begin{subfigure}[b]{.32\linewidth}
         \centering
         \includegraphics[width=\linewidth]{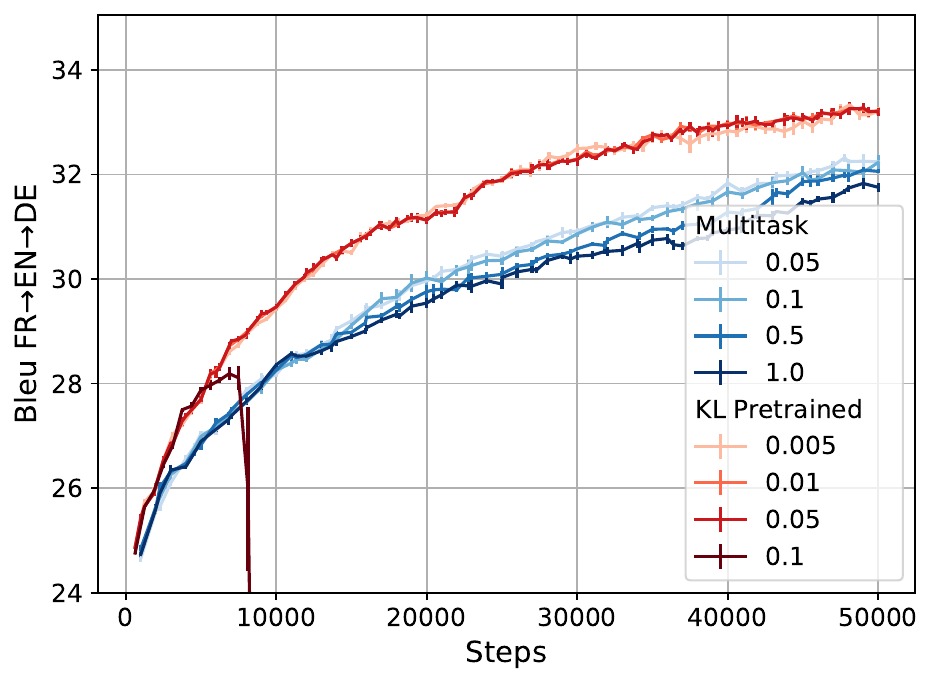}
         \caption{Task Performance}
     \end{subfigure}
     \begin{subfigure}[b]{.32\linewidth}
         \centering
         \includegraphics[width=\linewidth]{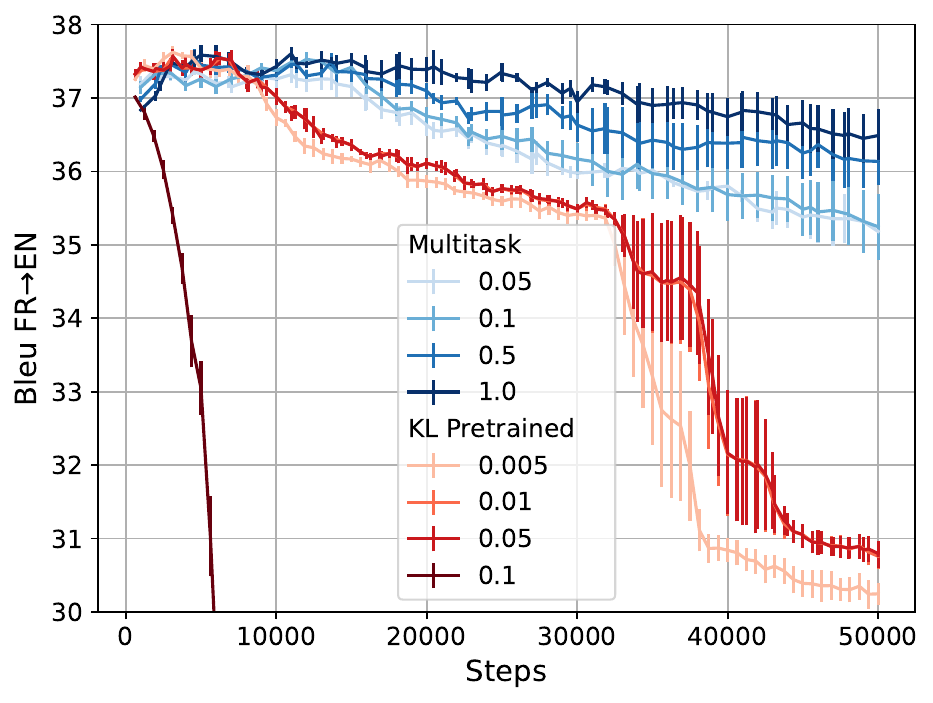}
         \caption{Language Drift}
     \end{subfigure}
     \begin{subfigure}[b]{.32\linewidth}
         \centering
         \includegraphics[width=\linewidth]{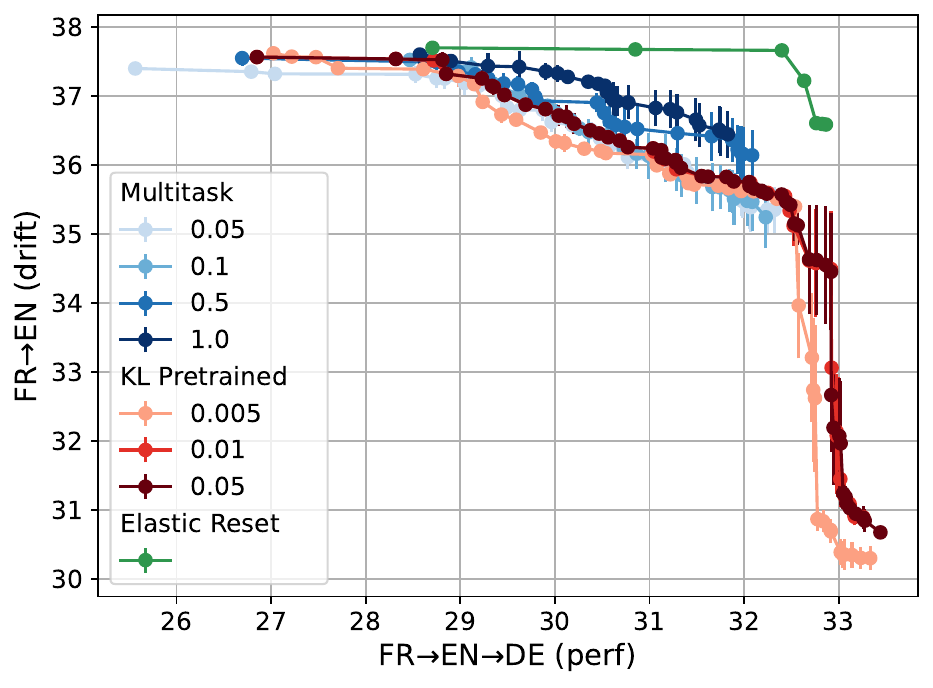}
         \caption{Pareto Frontier Graph}
     \end{subfigure}
    \caption{Ablation of KL and Multitask coefficients on Translation Game. We exclude KL 0.1 from the pareto graph since it fails. We plot the mean and show error bars for standard error over 5 seeds.}
    \label{fig:lang-klablation-multikl}
\end{figure}

As shown in Figure~\ref{fig:imdb-ablations}, Elastic Reset performance in relatively unaffected by the presence or absence of KL. In contrast, PPO and REINFORCE performance is known to be sensitive \citep{lu_countering_2020}. This raises the question of whether different coefficients of the KL loss would lead to different pareto frontiers. We find in Figure~\ref{fig:lang-klablation-multikl} that both Multitask and KL with Pretrained methods do have slightly different pareto frontiers. Choosing a 

\begin{figure}[ht]
     \centering
      \begin{subfigure}[b]{.32\linewidth}
         \centering
         \includegraphics[width=\linewidth]{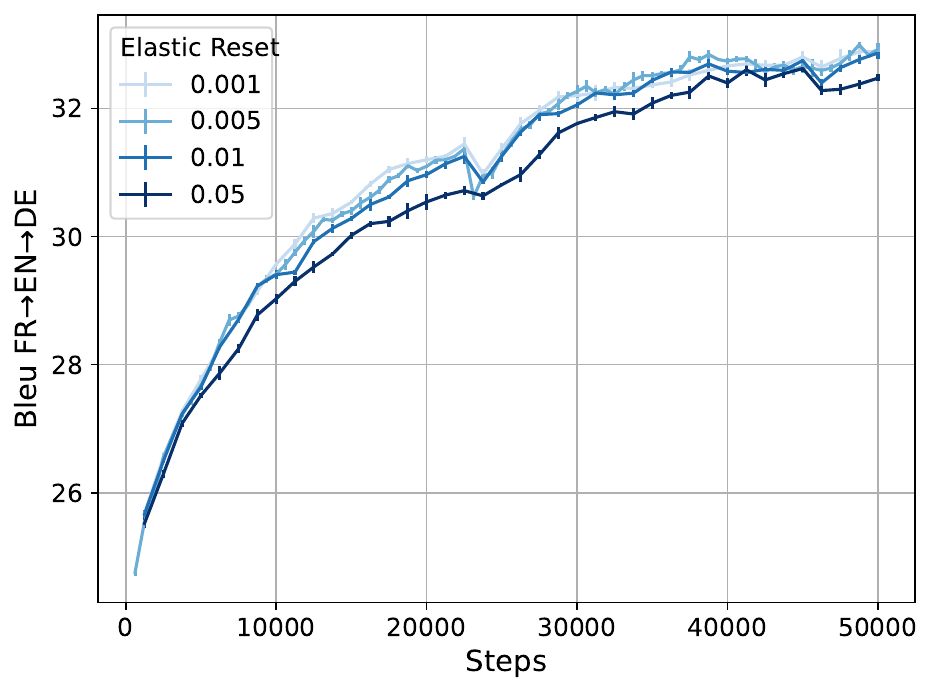}
         \caption{Task Performance}
     \end{subfigure}
     \begin{subfigure}[b]{.32\linewidth}
         \centering
         \includegraphics[width=\linewidth]{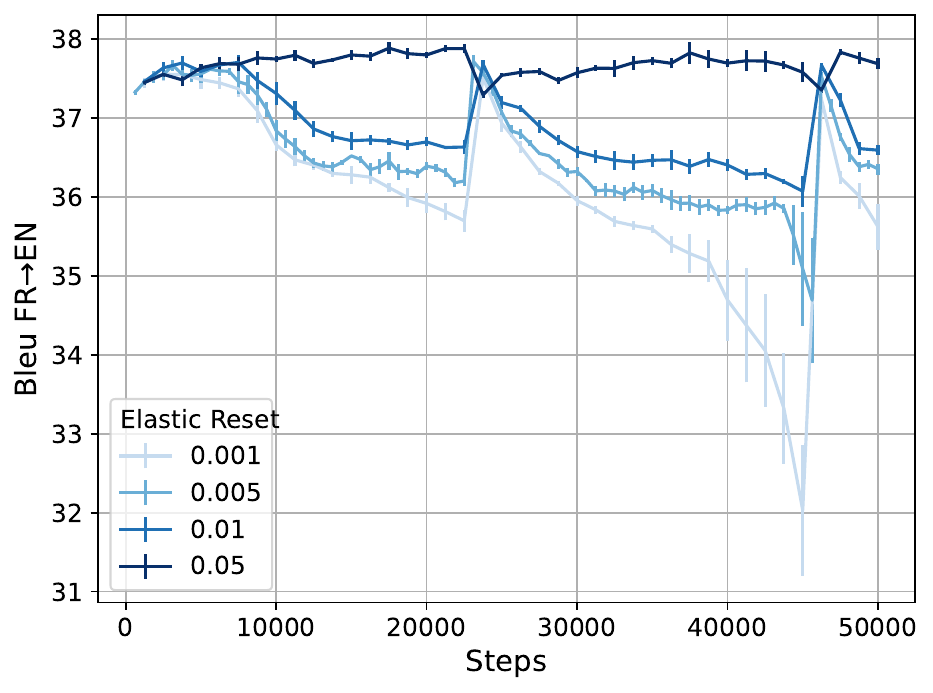}
         \caption{Language Drift}
     \end{subfigure}
     \begin{subfigure}[b]{.32\linewidth}
         \centering
         \includegraphics[width=\linewidth]{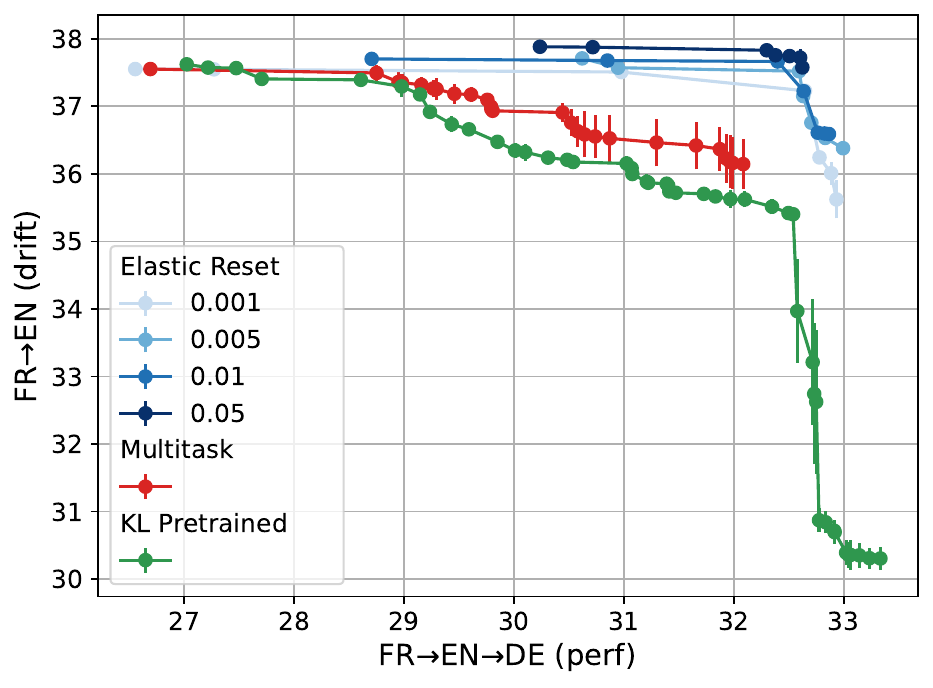}
         \caption{Pareto Frontier Graph}
     \end{subfigure}
    \caption{Ablation of KL coefficients for Elastic Reset on Translation Game. We plot the mean and show error bars for standard error over 5 seeds.}
    \label{fig:lang-klablation-emaset-extra}
\end{figure}

\begin{figure}[ht]
     \centering
      \begin{subfigure}[b]{.32\linewidth}
         \centering
         \includegraphics[width=\linewidth]{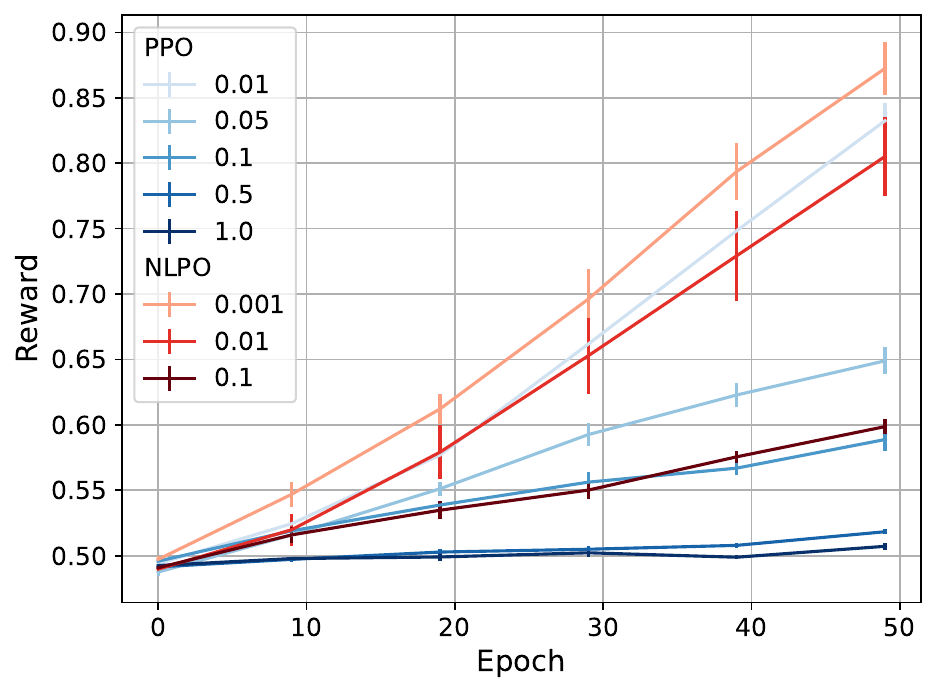}
         \caption{Task Performance}
     \end{subfigure}
     \begin{subfigure}[b]{.32\linewidth}
         \centering
         \includegraphics[width=\linewidth]{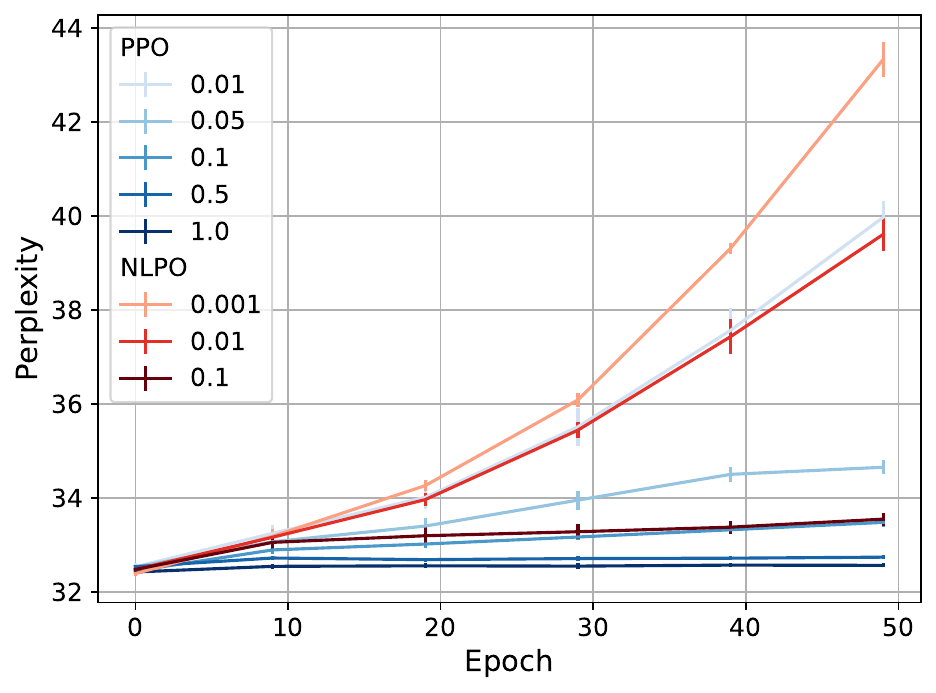}
         \caption{Language Drift}
     \end{subfigure}
     \begin{subfigure}[b]{.32\linewidth}
         \centering
         \includegraphics[width=\linewidth]{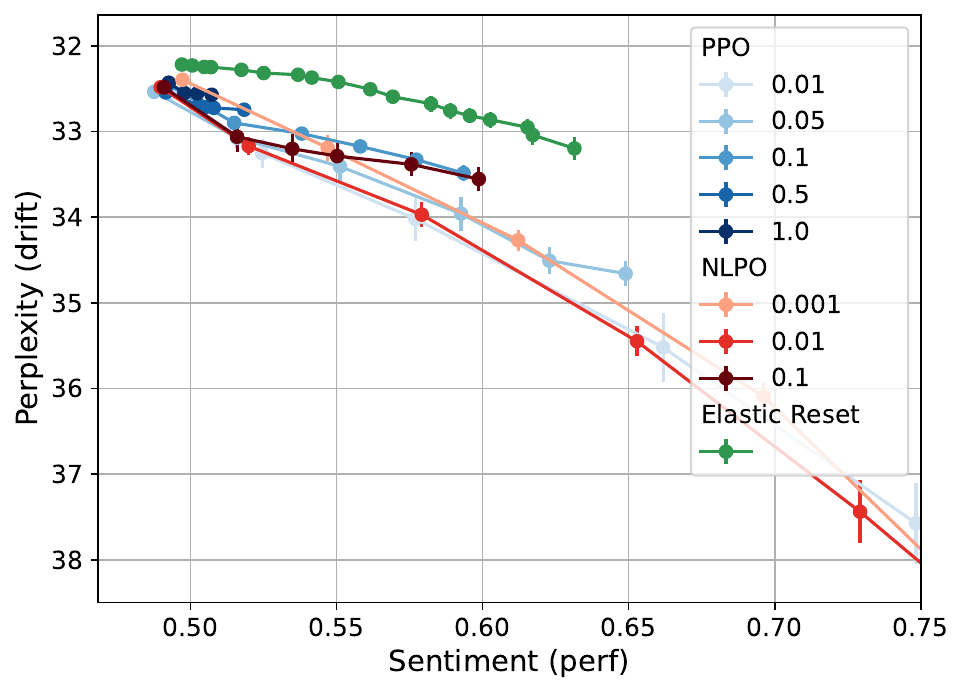}
         \caption{Pareto Frontier Graph}
     \end{subfigure}
    \caption{Ablation of PPO and NLPO coefficients on IMDB. We do more ablations of PPO since NLPO is always similar but worse than PPO. We plot the mean and show error bars for standard error over 5 seeds.}
    \label{fig:imdb-klablation-multikl}
\end{figure}

\begin{figure}[ht]
     \centering
      \begin{subfigure}[b]{.32\linewidth}
         \centering
         \includegraphics[width=\linewidth]{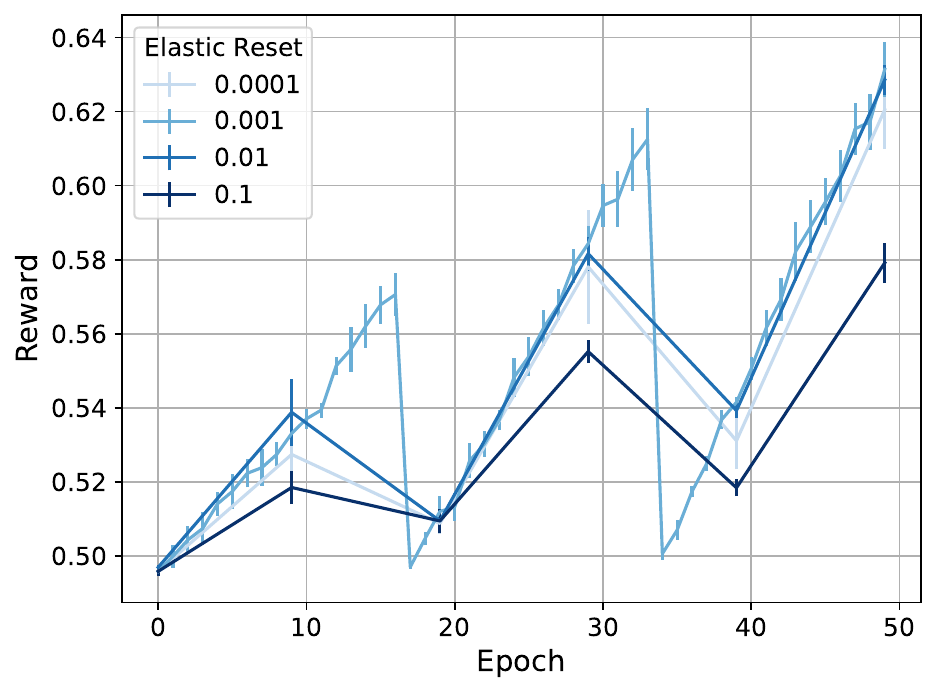}
         \caption{Task Performance}
     \end{subfigure}
     \begin{subfigure}[b]{.32\linewidth}
         \centering
         \includegraphics[width=\linewidth]{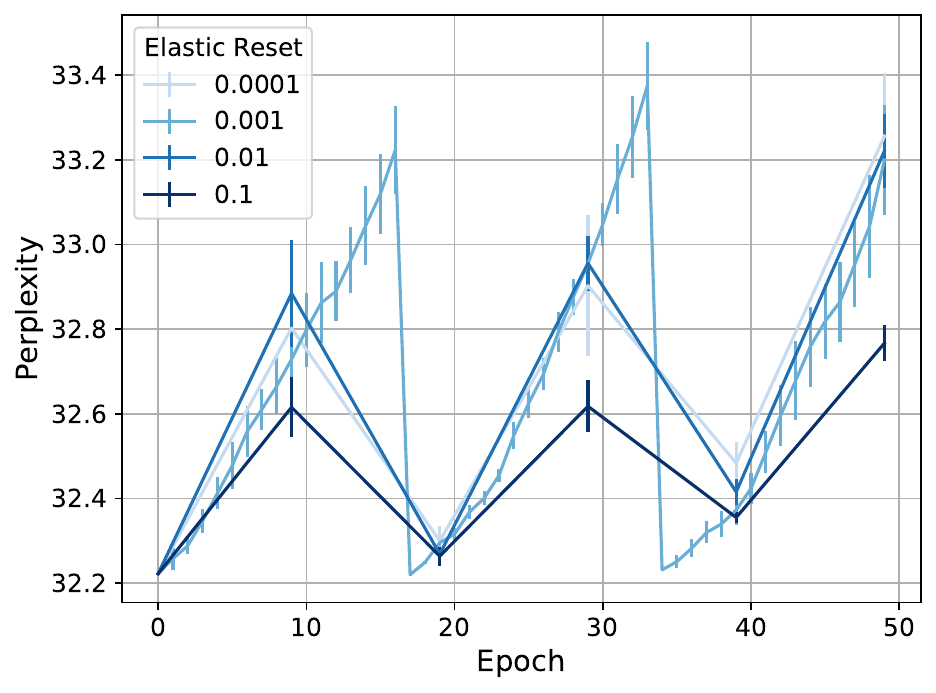}
         \caption{Language Drift}
     \end{subfigure}
     \begin{subfigure}[b]{.32\linewidth}
         \centering
         \includegraphics[width=\linewidth]{images/imdb_klablation_emaset_pareto.pdf}
         \caption{Pareto Frontier Graph}
     \end{subfigure}
    \caption{Ablation of KL coefficients for Elastic Reset on IMDB. We plot the mean and show error bars for standard error over 5 seeds.}
    \label{fig:imdb-klablation-emaset-app}
\end{figure}

\begin{figure}[ht]
     \centering
      \begin{subfigure}[b]{.32\linewidth}
         \centering
         \includegraphics[width=\linewidth]{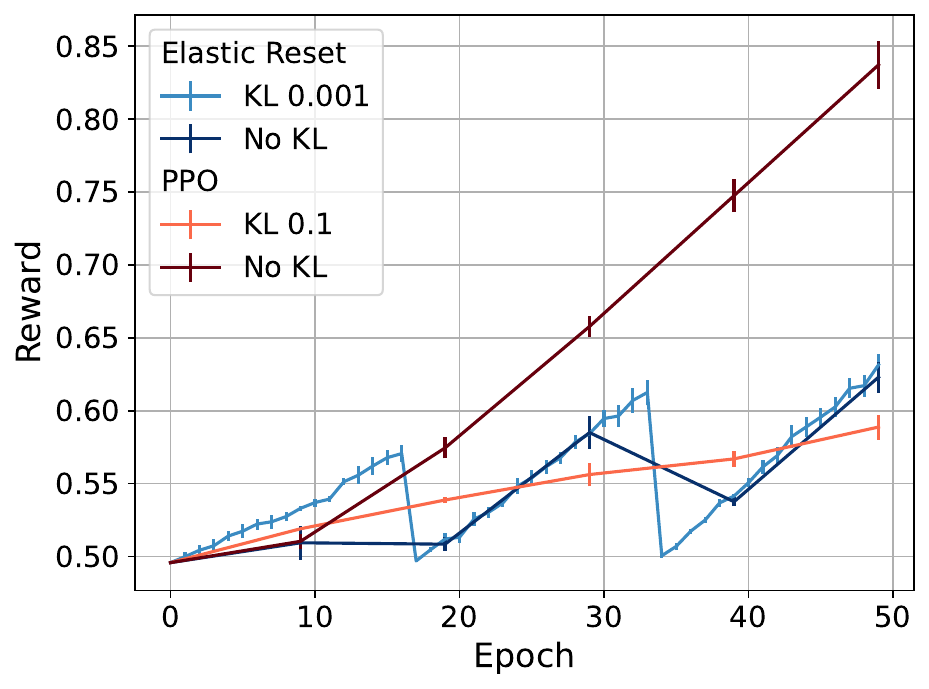}
         \caption{Task Performance}
     \end{subfigure}
     \begin{subfigure}[b]{.32\linewidth}
         \centering
         \includegraphics[width=\linewidth]{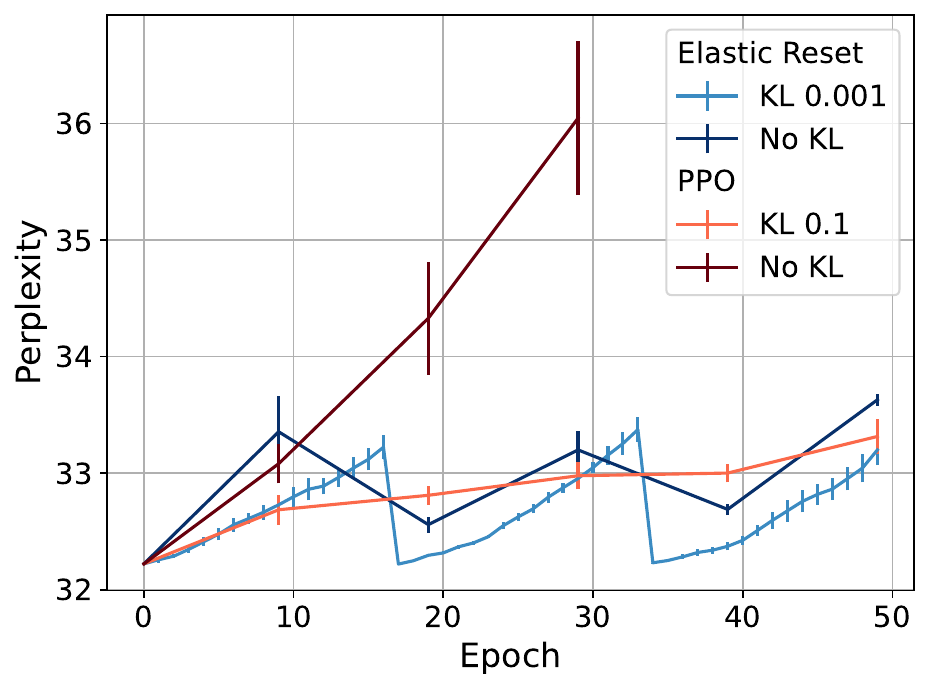}
         \caption{Language Drift}
     \end{subfigure}
     \begin{subfigure}[b]{.32\linewidth}
         \centering
         \includegraphics[width=\linewidth]{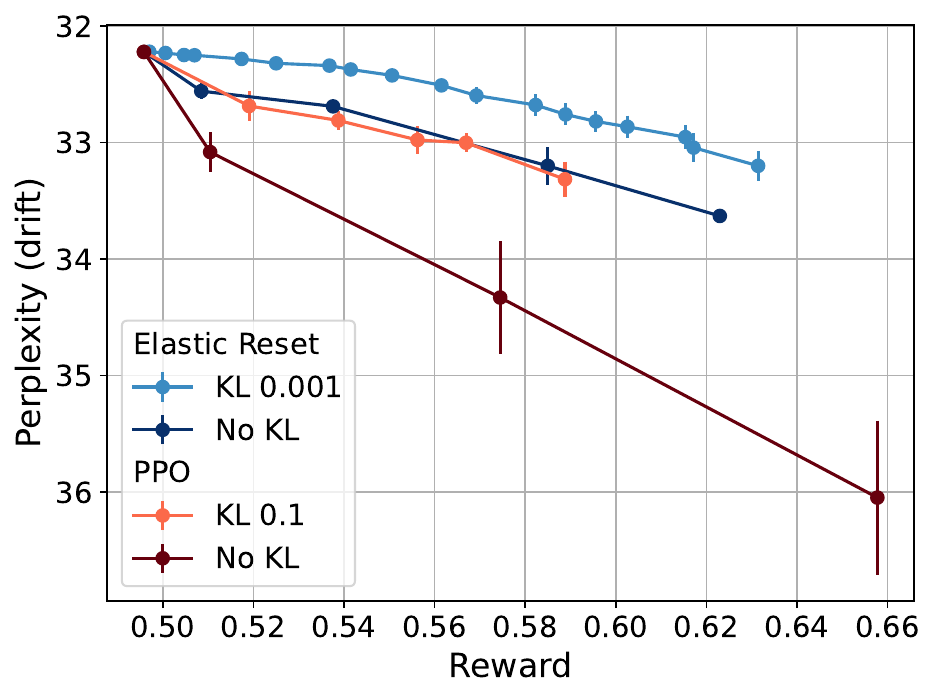}
         \caption{Pareto Frontier Graph}
     \end{subfigure}
    \caption{Ablation with/without KL penalty for PPO and Elastic Reset on IMDB. We plot the mean and show error bars for standard error over 5 seeds. For clarity, we cut off PPO without KL to four data points on the pareto graph since it diverges too much}
    \label{fig:imdb-kl0-emaset}
\end{figure}

\FloatBarrier
\subsection{Decay Rate}

\begin{figure}[ht]
     \centering
      \begin{subfigure}[b]{.32\linewidth}
         \centering
         \includegraphics[width=\linewidth]{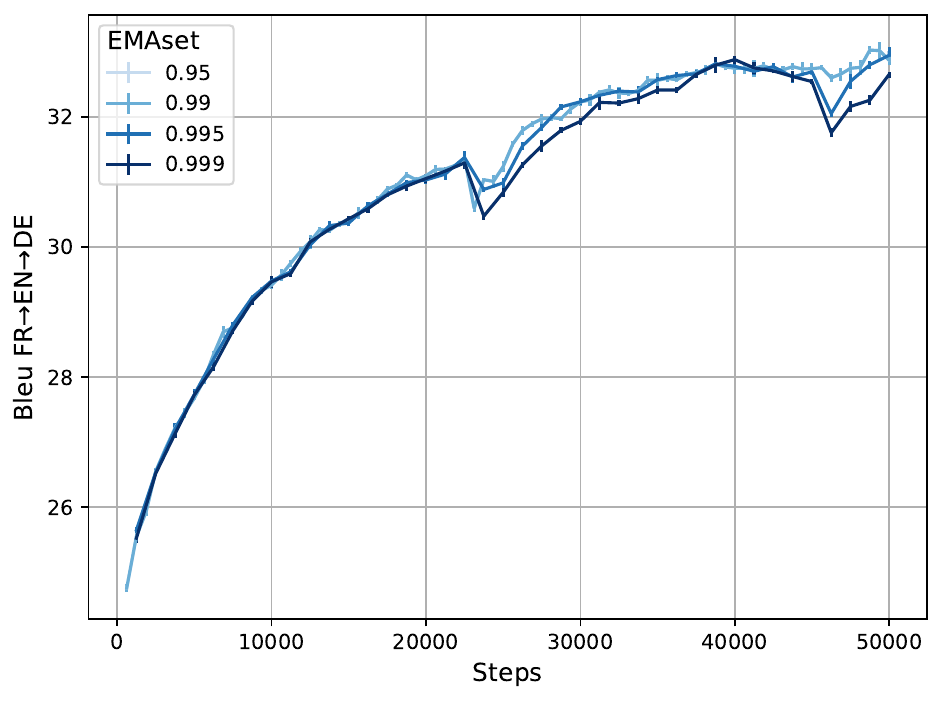}
         \caption{Task Performance}
     \end{subfigure}
     \begin{subfigure}[b]{.32\linewidth}
         \centering
         \includegraphics[width=\linewidth]{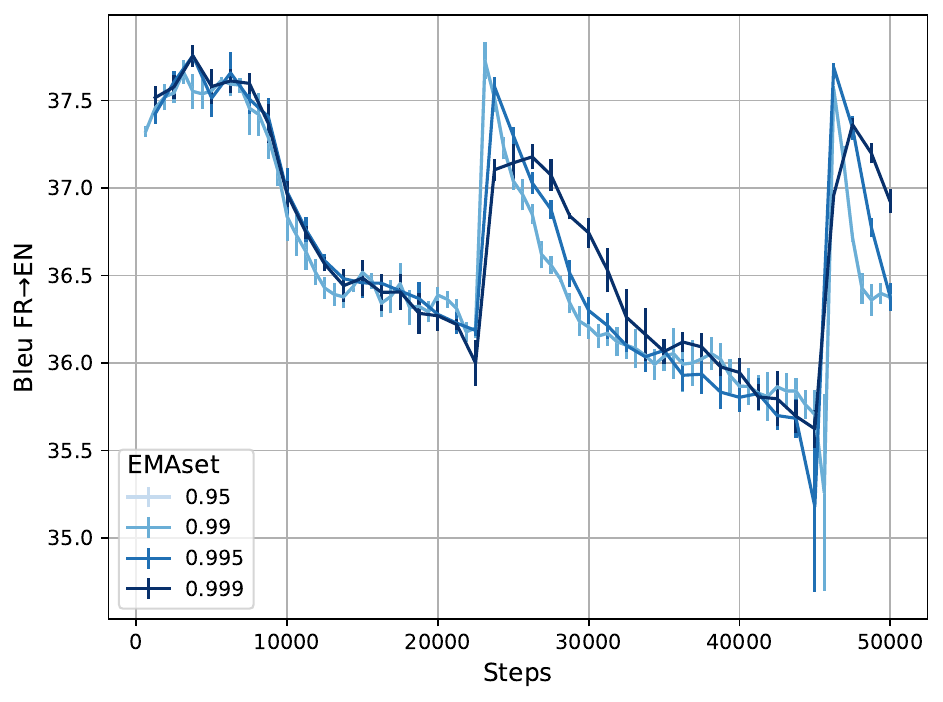}
         \caption{Language Drift}
     \end{subfigure}
     \begin{subfigure}[b]{.32\linewidth}
         \centering
         \includegraphics[width=\linewidth]{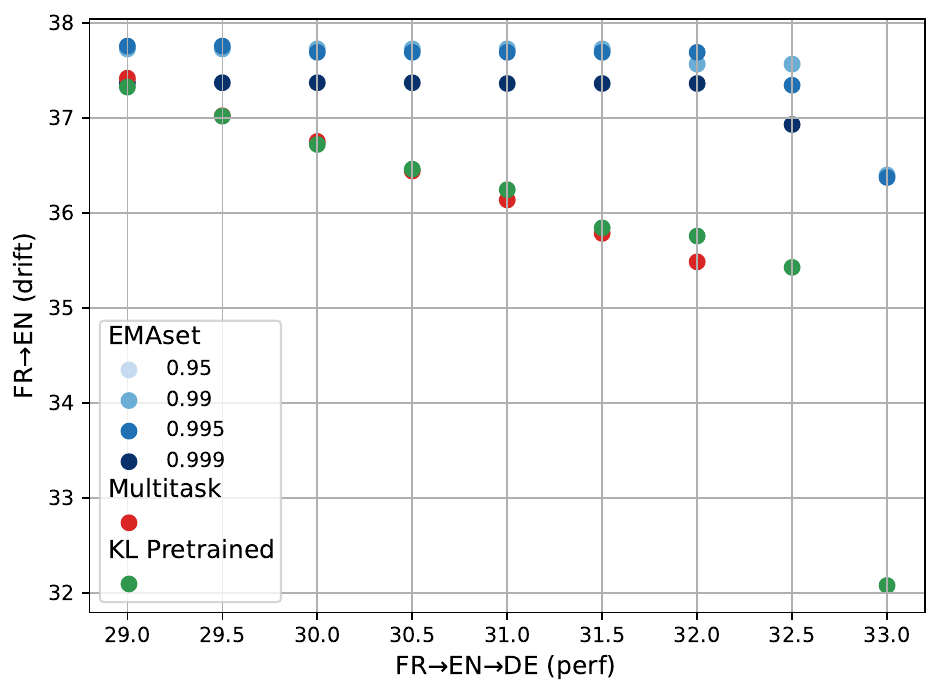}
         \caption{Pareto Frontier Graph}
     \end{subfigure}
    \caption{Ablation of decay coefficients for Elastic Reset on Translation Game. Original decay is 0.99. We plot the mean and show error bars for standard error over 5 seeds.}
    \label{fig:lang-decayablation-emaset}
\end{figure}

\begin{figure}[ht]
     \centering
      \begin{subfigure}[b]{.32\linewidth}
         \centering
         \includegraphics[width=\linewidth]{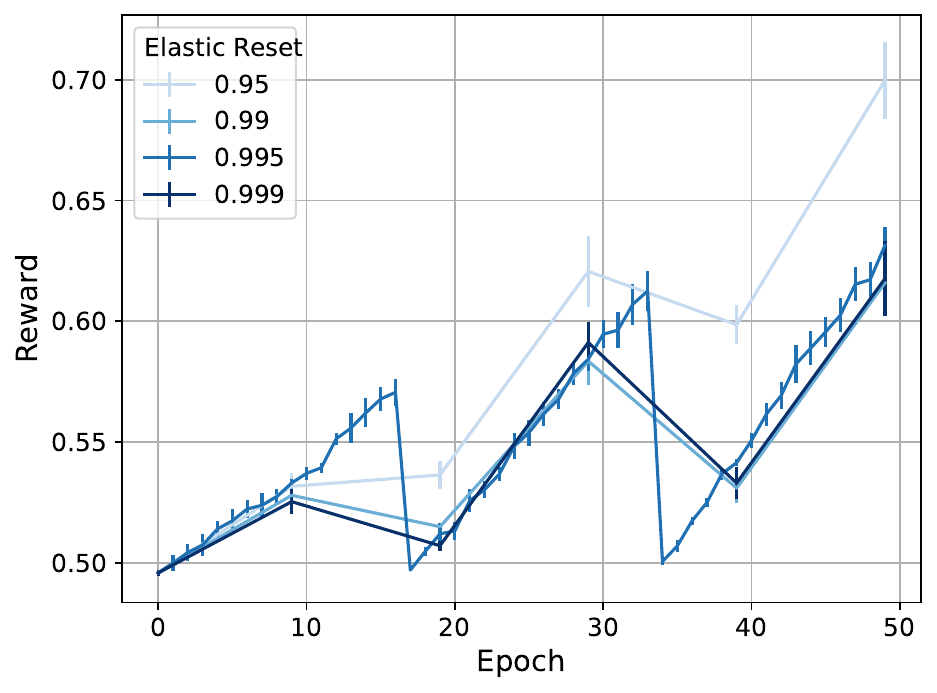}
         \caption{Task Performance}
     \end{subfigure}
     \begin{subfigure}[b]{.32\linewidth}
         \centering
         \includegraphics[width=\linewidth]{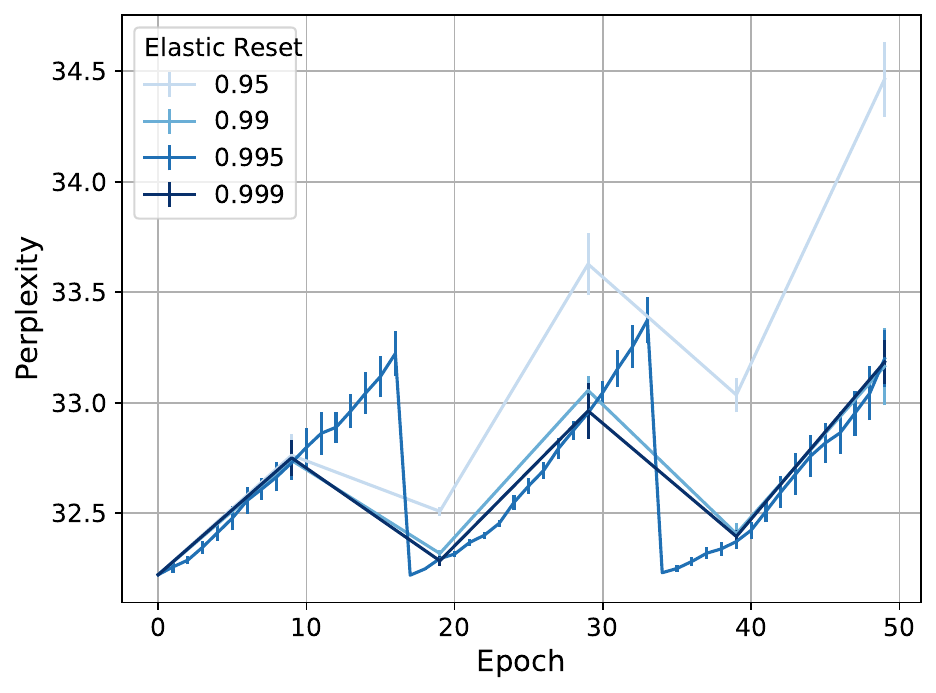}
         \caption{Language Drift}
     \end{subfigure}
     \begin{subfigure}[b]{.32\linewidth}
         \centering
         \includegraphics[width=\linewidth]{images/imdb_decay_emaset_pareto.pdf}
         \caption{Pareto Frontier Graph}
     \end{subfigure}
    \caption{Ablation of decay coefficient for Elastic Reset on IMDB. Original decay is 0.995. We plot the mean and show error bars for standard error over 5 seeds.}
    \label{fig:imdb-decayablation-emaset}
\end{figure}

\subsubsection{Reset Frequency}
\label{app:reset-freq}

\begin{figure}[ht]
     \centering
      \begin{subfigure}[b]{.32\linewidth}
         \centering
         \includegraphics[width=\linewidth]{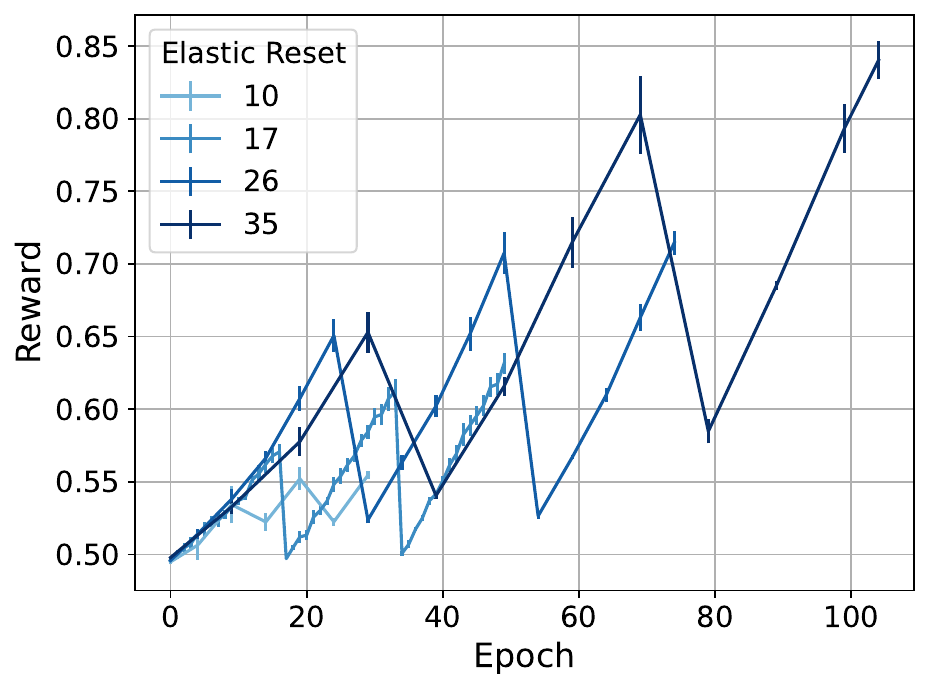}
         \caption{Task Performance}
     \end{subfigure}
     \begin{subfigure}[b]{.32\linewidth}
         \centering
         \includegraphics[width=\linewidth]{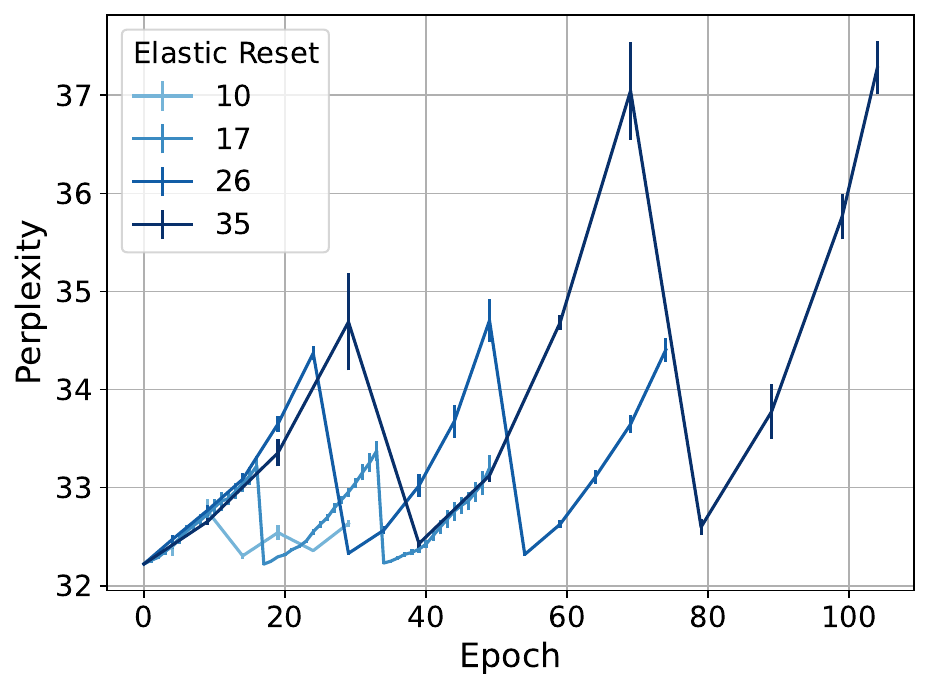}
         \caption{Language Drift}
     \end{subfigure}
     \begin{subfigure}[b]{.32\linewidth}
         \centering
         \includegraphics[width=\linewidth]{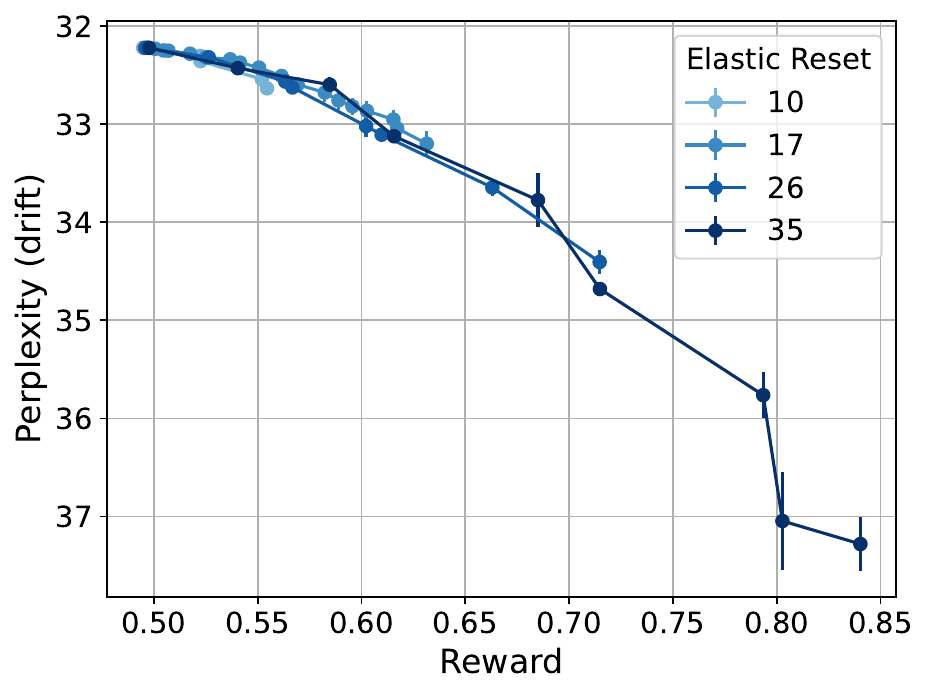}
         \caption{Pareto Frontier Graph}
     \end{subfigure}
    \caption{Ablation of reset frequency for Elastic Reset on IMDB. We plot the mean and show error bars for standard error over 5 seeds. Although 17 is optimal, results are relatively robust to choice of this hyperparameter}
    \label{fig:imdb-resetrate-emaset}
\end{figure}

\section{Explaining Elastic Reset}
Here we provide empirical intuition for why Elastic Reset is effective

\subsection{Explaining Resets: Value Function}
\label{app:explain-value}

Drift is a problem of noisy optimization, and Elastic Reset tackles this in two ways.

The first is by re-training with an improved value function. Though we begin with a sequence-level reward model, over training we learn a *token-level* value function. We believe that the value function greatly improves over time so early training may have exacerbated drift because it uses the early, sub-optimal value function. With each reset of the policy while maintaining the value function, we re-train with less noise and more direct gradients to alignment. Another way, both the reward model and optimal value function point towards high reward, but not in the same way. The reward model points towards high reward but not necessarily in the most direct way. Using the frozen reward model as our value function leads to equivalent reward but higher drift compared to using an optimal value function.

\begin{figure}[ht]
     \centering
      \begin{subfigure}[b]{.32\linewidth}
         \centering
         \includegraphics[width=\linewidth]{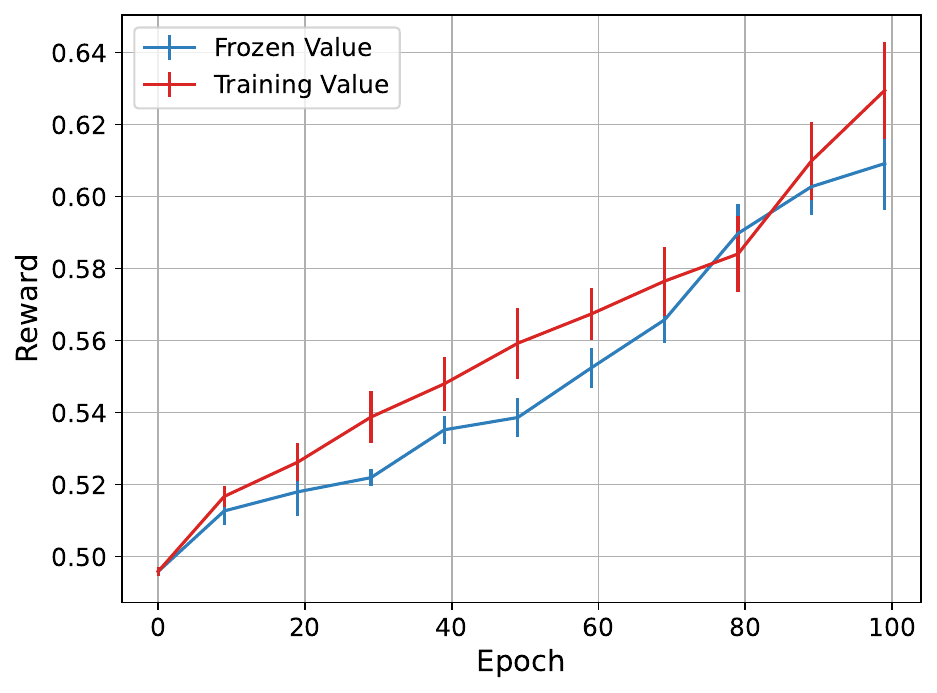}
         \caption{Task Performance}
     \end{subfigure}
     \begin{subfigure}[b]{.32\linewidth}
         \centering
         \includegraphics[width=\linewidth]{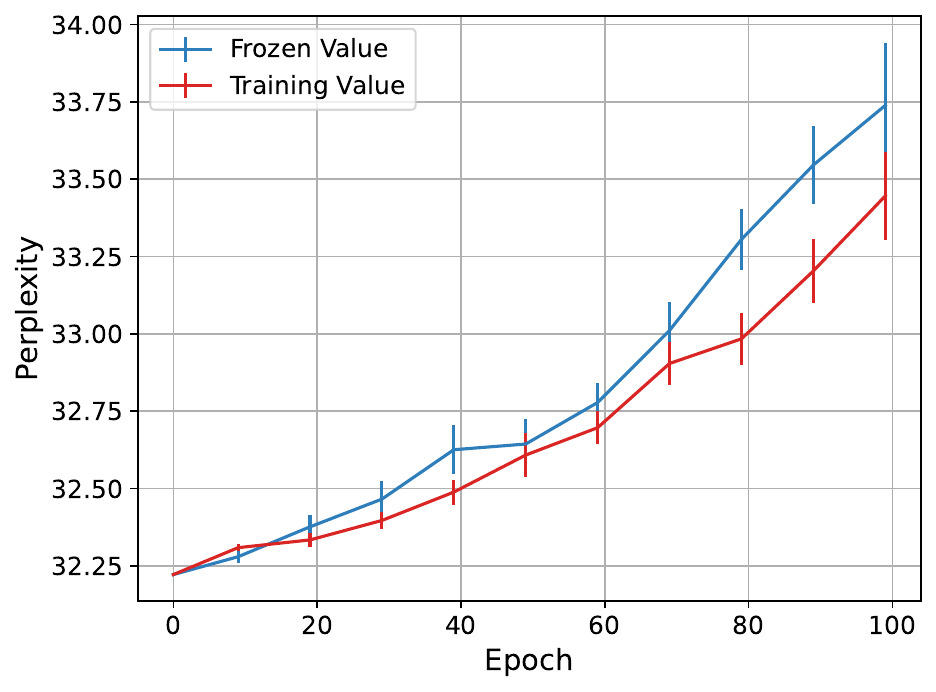}
         \caption{Language Drift}
     \end{subfigure}
     \begin{subfigure}[b]{.32\linewidth}
         \centering
         \includegraphics[width=\linewidth]{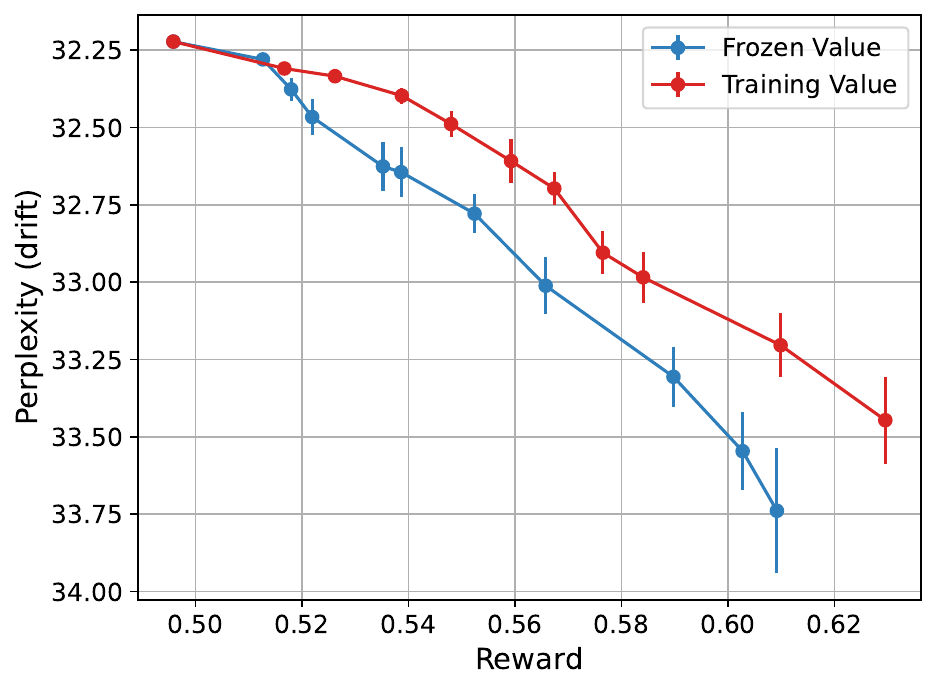}
         \caption{Pareto Frontier Graph}
     \end{subfigure}
    \caption{Ablation of learning vs frozen value function for PPO on IMDB.}
    \label{fig:imdb-frozenvalue}
\end{figure}

To demonstrate this phenomenon, we compared what happens if we don't train our value function at all but use a frozen reward model as our value function. Since our value function is GPT-2 but our reward model is DistilBERT, we first train a GPT-2 reward model similar to DistilBERT. We then compare regular PPO with a frozen value function i.e. just the sequence-level reward model vs a training value function. In Figure~\ref{fig:imdb-frozenvalue} we find that the learning value function is clearly better than the frozen one. They both reach the same reward but a more optimal value function leads to less drift for the same reward.

\subsection{Explaining Elastic Reset: The Benefit of EMA}
\label{app:explain-ema}

\begin{figure}[ht]
     \centering
      \begin{subfigure}[b]{.32\linewidth}
         \centering
         \includegraphics[width=\linewidth]{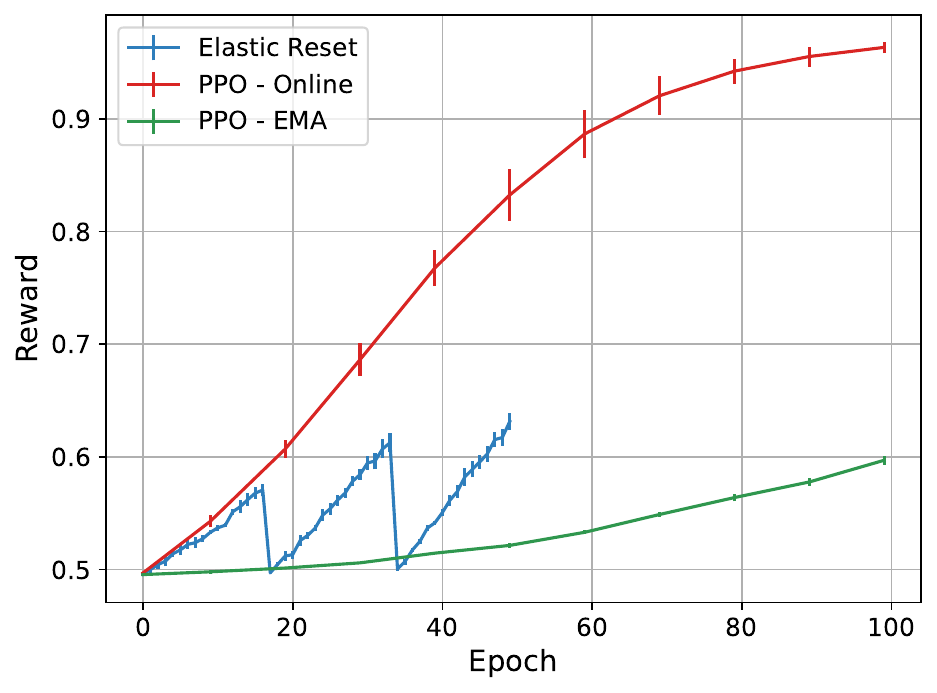}
         \caption{Task Performance}
     \end{subfigure}
     \begin{subfigure}[b]{.32\linewidth}
         \centering
         \includegraphics[width=\linewidth]{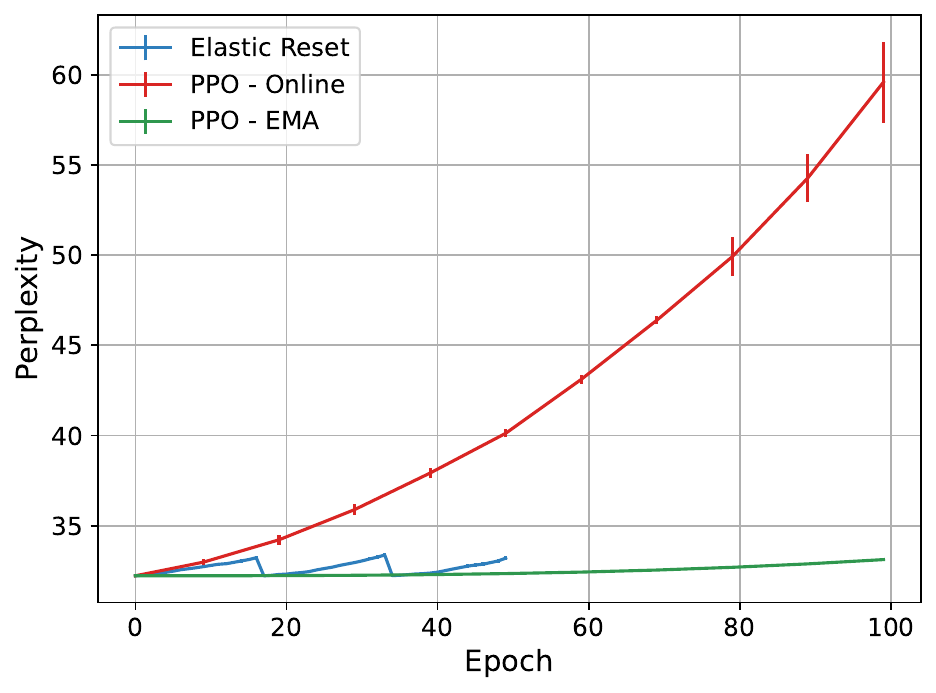}
         \caption{Language Drift}
     \end{subfigure}
     \begin{subfigure}[b]{.32\linewidth}
         \centering
         \includegraphics[width=\linewidth]{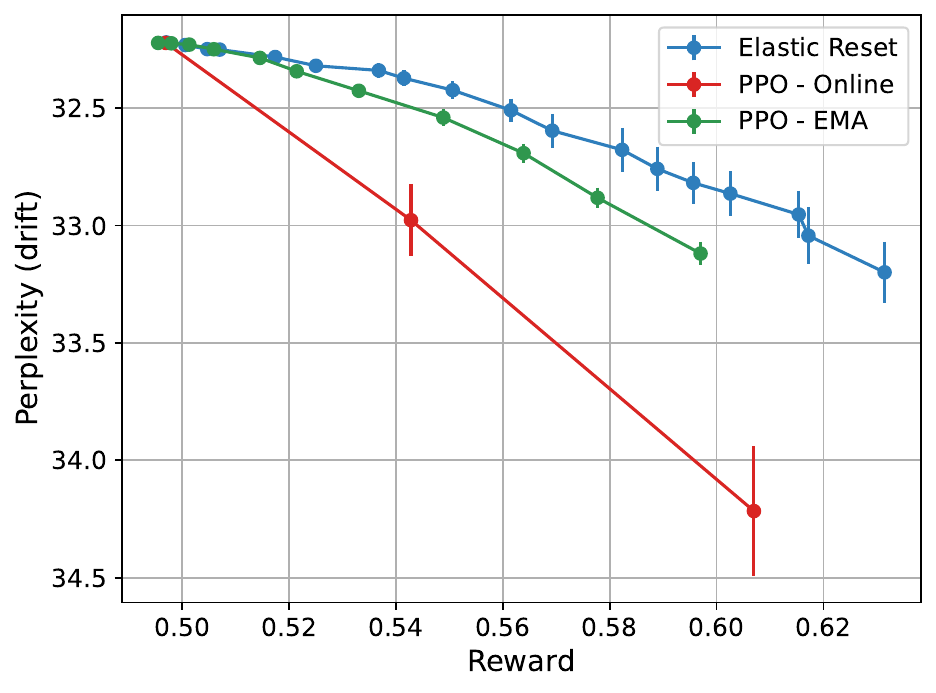}
         \caption{Pareto Frontier Graph}
     \end{subfigure}
    \caption{PPO trained on IMDB with a lower KL coefficient for 100 epochs, comparing its online network performance vs EMA model performance. We cut off PPO - online to just three data points in the pareto graph for clarity and visual scale.}
    \label{fig:imdb-ema}
\end{figure}

The second is by smoothing optimization with an EMA. Work in other fields e.g. DINO in SSL, has leveraged EMA for stability and improved generalization. By resetting to an EMA and resetting our EMA, we further smooth the gradients to alignment.  To demonstrate, we run PPO as usual but keep track of its EMA, without resetting to it. We then plot the accuracy of the online and EMA networks over training. To make the effect larger, we use a smaller KL $\beta = 0.01$ and train our PPO for 100 epochs and plot results in Figure~\ref{fig:imdb-ema}. We find that just keeping the EMA model is an effective way to improve the task / drift tradeoff. But we also see that it is noticeably slower than our method. PPO - Online needs to reach nearly the maximum possible reward for its EMA to improve only slightly on performance. 

% \FloatBarrier

% \newgeometry{}

% \section*{Rebuttal Additional PDF}
% \setcounter{figure}{0}

% \begin{figure}[ht]
%      \centering
%       \begin{subfigure}[b]{.25\linewidth}
%          \centering
%          \includegraphics[width=\linewidth]{images/stackllama_reward_smooth.pdf}
%          \caption{Reward}
%      \end{subfigure}
%      \begin{subfigure}[b]{.25\linewidth}
%          \centering
%          \includegraphics[width=\linewidth]{images/stackllama_kl_smooth.pdf}
%          \caption{KL from Initial}
%      \end{subfigure}
%     \caption{Rolling Window Mean (window = 10) version of the StackLLaMA results}
% \end{figure}

\end{document}